\documentclass[11pt]{article}

\usepackage[final]{acl}
\usepackage{xcolor,booktabs,makecell,siunitx}
\usepackage{booktabs,tabularx,ragged2e}
\newcolumntype{Y}{>{\RaggedRight\arraybackslash}X}

\usepackage{booktabs,tabularx,ragged2e,makecell}
\newcolumntype{L}[1]{>{\RaggedRight\arraybackslash}p{#1}}
\newcolumntype{Y}{>{\RaggedRight\arraybackslash}X}
\usepackage{amsmath}
\usepackage{enumitem}
\usepackage{algorithm}
\usepackage[noend]{algpseudocode} 
\usepackage{amssymb}  
\usepackage{todonotes}

\usepackage{comment}
\usepackage{times}
\usepackage{latexsym}

\usepackage[T1]{fontenc}

\usepackage[utf8]{inputenc}

\usepackage{microtype}

\usepackage{inconsolata}

\usepackage{booktabs}
\usepackage{graphicx}
\usepackage{float}

\title{Do Image–Text Metrics Respect Semantic Invariances?}

\author{
  Amit Agarwal\textsuperscript{}\quad Hitesh Laxmichand Patel \quad Meizhu Liu \quad Jyotika Singh \\
  \textbf{Karan Dua} \quad \textbf{Hansa Meghwani} \quad \textbf{Matthew Rowe} \quad \textbf{Michael Avendi} \\
  \textbf{Yassi Abbasi} \quad \textbf{Tao Sheng} \quad \textbf{Sujith Ravi} \quad \textbf{Dan Roth} \\
  Oracle AI \\
  \small{
   \textbf{Correspondence:} \href{mailto:amit.h.agarwal@oracle.com}{amit.h.agarwal@oracle.com}
 }
}

\begin{document}
\maketitle

\begin{abstract}
Reference-free image-to-text evaluators are now standard for scoring image-caption alignment, yet it is unclear whether they respect semantic invariances. We present an invariance probe on 
five popular evaluators (CLIPScore, PAC-S, UMIC, FLEUR, and a deterministic LLM judge) under semantics-preserving perturbations along three axes- spatial (flips, context-preserving repositioning, light rotations), object (scale, category), and socio-linguistic framing (cultural/economic adjectives with neutral and length-matched controls). Across curated slices of three detection datasets and three caption evaluation suites, we find consistent non-semantic sensitivities, where benign spatial edits and simple phrasing changes shift scores by \(\approx\)6-9\% on average, and for systems separated by just 0.7\%, these shifts can cause ranking flips in upto \(\sim\)37\% of cases, particularly under spatial changes. A small human study also supports this finding and confirms that annotators generally judge perturbed pairs as equally correct, so these shifts reflect metric behavior rather than semantic change. We further propose invariance-calibrated scoring, a post-hoc adjustment that roughly halves median absolute sensitivity while retaining correlation with learned caption evaluators.
\end{abstract}

\section{Introduction}
Reference-free captioning evaluators are now standard in multimodal research in which they scale without human references and often show strong correlations with human judgments (e.g., \textsc{CLIPScore}~\citep{hessel2021clipscore}, \textsc{PAC-S}~\citep{Sarto2023PositiveAugmentedCL}). Yet such aggregate correlations can mask how evaluators behave under semantics-preserving changes, which are alterations to the layout, object factors, or wording that do not affect the core meaning of the content. A metric may appear stable on average while responding systematically to vertical flips, repositioning that preserves the meaning of the scene, or socially marked adjectives that do not alter the depicted content. When systems are separated by sub-percent gaps, these sensitivities can invert rankings and confound fairness conclusions. In deployed NLP stacks, metric outputs drive downstream decisions ranging from model selection and reward shaping to retrieval ranking and the scoring of natural-language renditions of structured outputs~\citep{nlprealworld, hardnegmining, llmnarratetables}, so evaluator reliability is a practical concern beyond any single benchmark.

We take a \emph{metric-centric invariance audit}: rather than correlating to references or annotators, we maintain a fixed meaning and apply controlled probes across three families (\textit{spatial}, \textit{object}, and \textit{socio-linguistic}) to test whether scores remain stable. We instantiate this on three image sources (COCO~\cite{COCO}, OpenImages~\cite{OpenImages}, Objects365~\cite{Objects365}) and five evaluators spanning embedding similarity and learned caption assessment (\textsc{CLIPScore}, \textsc{PAC-S}, \textsc{UMIC}~\cite{UMIC}, \textsc{FLEUR}~\cite{lee-etal-2024-fleur}, and one deterministic judge). A human validation study confirms that, for almost all paired items, annotators judge both versions acceptable and equally good, so systematic score changes reflect evaluator behavior rather than semantic drift.


We introduce \emph{invariance-calibrated scoring}, a post-hoc adjustment that subtracts per-prompt nuisance sensitivity estimated from invariance families. Under a tight correlation-preservation constraint with learned caption evaluators, it roughly halves median absolute sensitivity on average (largest on spatial probes) while retaining evaluator utility and reducing ranking-flip risk.
The contribution is diagnostic, not a claim that one metric objective is correct. Whether evaluators are used for alignment or typicality, the magnitude and asymmetry of the shifts we document warrant disclosure and the opt-in mitigation we propose.
Our contributions can be summarized as: 

\begin{itemize}
  \item A unit-test framework that audits reference-free captioning evaluators for \textit{semantic invariance} across spatial, object, and socio-linguistic families.
  \item A cross-source, cross-metric study revealing consistent, practically meaningful sensitivities ($\approx6$-$9\%$ under benign spatial changes) and linking them to leaderboard instability via an intuitive flip-risk functional.
  \item A practical mitigation, \emph{invariance-calibrated scoring}, that reduces non-semantic sensitivity without retraining.
\end{itemize}


\section{Related Work}

Reference-free metrics have reshaped evaluation for vision-language models, whose dataset and application landscape has expanded rapidly~\citep{mmsurvey}, offering scalable alternatives to reference-based measures such as CIDEr~\cite{CIDEr} and SPICE~\cite{SPICE}. Among them, {CLIPScore} has been widely adopted, as it computes the alignment between image and caption using pre-trained CLIP embeddings and often correlates better with human judgments than traditional metrics. This has led to its use in selection and reward pipelines~\citep{cho2023finegrainedimagecaptioningclip} and analyses of grounding behavior~\citep{barraco2022unreasonable}. To mitigate documented weaknesses, {PAC-S} and PAC-S++~\cite{Sarto2023PositiveAugmentedCL,sarto2025positiveaugmentedcontrastivelearningvisionandlanguage} improve robustness to redundancy and noise. Complementary families such as HICE-S~\citep{Zeng_2024} and ensembling approaches like ECO~\citep{jeong2024technicalreportnicechallenge} and BRIDGE~\citep{sarto2023positiveaugmentedcontrastivelearningimage} combine signals or refine representations to stabilize alignment. Surveys map the space of vision–language evaluation~\citep{Zhang2024VisionLanguageSurvey}, while critiques examine when automatic scores capture meaning rather than surface regularities~\citep{ross2024makesgoodmetricevaluating}. Most work reports aggregate correlations, leaving unclear how metrics behave under semantics-preserving perturbations. Evidence suggests these failures arise from limitations of CLIP-like embeddings and transformer reasoning ~\citep{barriersreasoning}, motivating interpretable alternatives such as DCSMs ~\citep{kang2025clip}. Complementary lines of work benchmark model-side robustness of vision--language systems under targeted perturbations and context shifts, and propose fine-grained scores for multimodal reasoning quality~\citep{mvtamperbench, pcri, rci}; our audit instead targets the evaluator itself.

A growing body of work also interrogates fairness, accessibility, and framing. Studies probe whether reference-free metrics reflect linguistic intent or user needs~\citep{ahmadi2024examinationrobustnessreferencefreeimage,kasai2022transparenthumanevaluationimage}, highlight accessibility gaps~\citep{zur2024updatingclippreferdescriptions}, and propose fairness-oriented designs such as FLEUR~\citep{lee-etal-2024-fleur}. Parallel efforts also curate multicultural vision--language resources to counter Eurocentric defaults in benchmarks and evaluation, for instance SEA-VL for Southeast Asia~\citep{seavl}. Distinct from prior work, we present a \emph{metric-centric} unit-test audit that keeps core meaning fixed while varying image structure (e.g., spatial position, object scale) and language (e.g., framing, word choice), yielding actionable diagnostics for bias-aware design and reporting. A parallel line of work studies LVLMs as judges for vision--language
tasks and examines their reliability and biases
\citep{zhang2023gpt,chen2024mllm,hwang2025fooling};
our deterministic judge is representative of this family, and our
framework treats any such judge as an additional scoring function
to be audited rather than as a gold standard.


\begin{figure*}[h]
    \centering
    \includegraphics[width=1\linewidth]{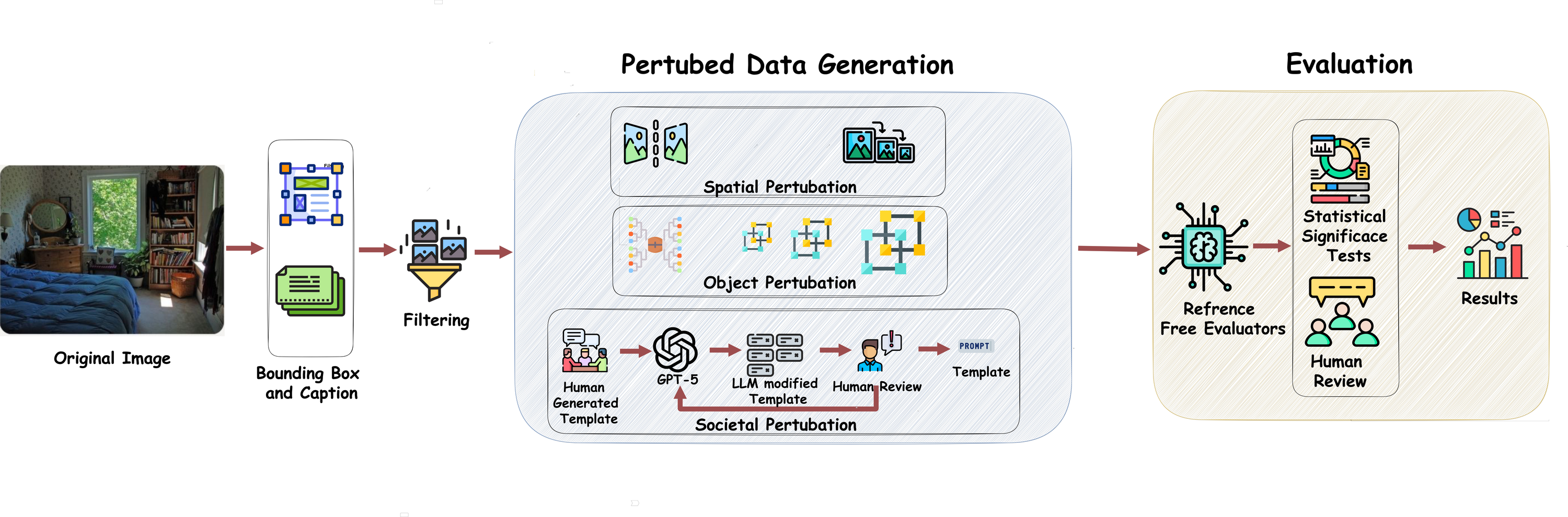}
    \caption{Pipeline Overview. Starting from curated single-object image--caption pairs, we construct matched spatial, object, and socio-linguistic variants and quantify reference-free evaluators sensitivity with paired comparisons.}
    \label{fig:evaluation_framework}
    \vspace{-1.5em}
\end{figure*}

\section{Methodology}
\label{sec:method}

We propose an \emph{invariance audit} for captioning-oriented, reference-free vision-language evaluators where the principle is simple, if an image-caption pair's meaning is unchanged, a reliable evaluator should be stable. Rather than relying on aggregate correlations, we apply controlled, semantics-preserving perturbations along spatial, object, and socio-linguistic axes while holding semantics fixed. 



We adopt an \emph{evaluation-centric} notion of semantic invariance, in which a perturbation is semantics-preserving when human annotators judge the caption-image relation as equally acceptable for caption-quality scoring. Under this definition, an evaluator used for alignment or caption quality should not materially shift. The validation study bears this out, with both versions majority-acceptable in $97.3\%$ of paired items and majority-tied preferences in $96.6\%$ (Appendix~\ref{app:human_validation}). The scope is deliberate, we do not claim that sensitivity to typicality is inherently a defect of a representation. But when these metrics are deployed for model selection, leaderboards, or reward pipelines, undisclosed sensitivity to nuisance factors produces measurement volatility and ranking instability, regardless of whether one ultimately reads it as a bug or a feature of the encoder.

\paragraph{Metrics under audit.}
We study five captioning-centric, reference-free evaluators spanning embedding similarity (\textsc{CLIPScore}, \textsc{PAC-S}) and learned caption quality (\textsc{UMIC}, \textsc{FLEUR}), plus one deterministic LLM-judge variant with a fixed rubric to triangulate with judge-style practice.\footnote{Reference-based metrics such as CIDEr and SPICE measure agreement with references and are not the focus of this invariance audit.} The framework is evaluator-agnostic where adding a new scoring function $S(x,c)$, for instance an open-source LVLM-as-judge, only requires implementing a scoring interface.

\subsection{Datasets and Curation}
\label{sec:datasets}

We audit three image sources: \textbf{MS-COCO 2017 val}, \textbf{OpenImages V7 val}, and \textbf{Objects365 val}, and we replicate key probes 
on three caption-evaluation suites (\textbf{Flickr8k-CF}, \textbf{Pascal-50S}, and \textbf{COMPOSITE}) for external validity. For the caption suites we follow the official image pools to avoid domain shift. Across all sources, we use a shared curation protocol and harmonize labels to a single taxonomy.

\paragraph{Single-object slice.}
From each large-scale source we select images with one prominent instance and high-quality boxes:
(i) a single dominant object (with a maximum Intersection over Union (IoU) overlap of < 0.1 with other objects, where IoU measures the ratio of intersection area to the union area between two bounding boxes);
(ii) clear visibility;
(iii) compatibility with planned transforms (no truncation after repositioning/rotation);
(iv) box area sufficient to stratify size.
We bin normalized coverage (box area / object area) into seven thresholds: 0-10, 10-20, 20-35, 35-50, 50-70, 70-90, 90-100\%.
Per-source balance tables (category $\times$ size bin) appear in Appendix~\ref{app:balance}.
This controlled single-object, templated-caption regime is a deliberate choice to maximize internal validity such that each perturbation targets a specific spatial, object, or socio-linguistic factor while the depicted content remains fixed. External validity under this design is addressed in Section~\ref{sec:res-robust} (caption-evaluation suites, naturally occurring captions, and a multi-object probe), across which the main conclusions persist.

\paragraph{Category harmonization.}
We map dataset 
labels into a shared taxonomy-
\emph{person}, \emph{animal}, \emph{vehicle}, \emph{furniture}, \emph{kitchen}, \emph{sports}, \emph{electronics}, \emph{indoor}, \emph{outdoor}.
Ambiguous classes are resolved by majority usage; dataset composition and mapping details are reported in Appendix~\ref{app:balance}.

\paragraph{Caption templates and lexical families.}
Captions are programmatically generated from human-designed templates to preserve semantics while varying framing. Controlled synthesis of this kind, where content-bearing factors are parameterized while surrounding structure is held fixed, is increasingly used for training and evaluation data in multimodal pipelines~\citep{flexdoc, docsynth}.
We use simple declarative forms such as
\emph{``There is a [object].''} and
\emph{``There is a [adjective] [object].''};
minimal spatial descriptors are added only when needed for disambiguation so that wording does not encode the downstream transforms.
Socio-linguistic modifiers cover \emph{cultural}, \emph{economic}, \emph{gender}, and \emph{emotion} families, as well as a small set of occupational and socio-political descriptors (e.g., \emph{local}/\emph{foreign}, \emph{immigrant}/\emph{citizen}) used in targeted probes. Each family includes \emph{neutral} baselines (e.g., \emph{typical}, \emph{plain}) and \emph{length-matched} variants to control for phrasing length and syntax. The length-matching process ensures that each variant has a similar number of characters or tokens, preventing any bias due to differences in length. For a subset of analyses, we also form simple intersectional combinations (e.g., cultural modifiers applied to person vs.\ non-person objects) to probe whether sensitivities depend on the underlying category. Further details on screening and lexicons are detailed in the Appendix~\ref{app:lexicons}).



\subsection{Perturbation Axes}
\label{sec:axes}

We construct semantics-preserving perturbations to isolate non-semantic sensitivities while keeping the described content fixed. Each probe is evaluated as a paired contrast against the unperturbed version.
Perturbations are grouped into three families - 

\textbf{Spatial.}
We apply vertical and horizontal flips, \emph{context-preserving repositioning}, and light in-plane rotations ($\pm 10^\circ$).
For repositioning, the segmented dominant object is translated within the \emph{original background} at constant scale to four anchors (TL, TR, BL, BR); translations that would collide with boundaries or occlude salient regions are resampled.
We also include a \emph{Gaussian-blur control} ($\sigma\!\in\!\{1.0,2.0\}$) on the original image to decouple size from texture/detail loss. To rule out compositing artifacts, we quantify background-change and boundary-seam indicators for repositioned images and verify that 
the findings persist after filtering artifact-heavy cases (Appendix~\ref{app:repositioning_diagnostics}).

\textbf{Object.}
We analyze scores across the seven coverage bins and the harmonized taxonomy while holding captions fixed, probing how object scale and category influence metric behavior independent of wording.

\textbf{Societal.}
We replace neutral captions with culturally, economically, and socially marked variants (e.g., \emph{American}/\emph{European}, \emph{cheap}/\emph{expensive}) plus gender and emotion adjectives, chosen to alter socio-linguistic framing while leaving scene semantics unchanged (e.g., \emph{African bed} vs.\ \emph{American bed} for the same pictured bed). All comparisons include \emph{neutral} and \emph{length-matched} controls so that differences reflect framing rather than phrasing length. We analyze aggregate effects and stratified slices (person vs.\ non-person; Appendix~\ref{app:societal_appendix}), and complement templated probes with naturally occurring captions (Appendix~\ref{app:real_captions}); perturbation-axis details are in Appendix~\ref{app:perturb_impl}.

\paragraph{Human validation of semantics preservation.}
We validate that our perturbations are semantics-preserving for humans via a small study on $N_\text{human}=480$ paired items from the curated slice (3 annotators/item; Appendix~\ref{app:human_validation}), each comparing two versions of the same example (image perturbation with a shared caption, or caption perturbation with a shared image). A majority judges \emph{both} versions acceptable in 467/480 items (97.3\%), and the majority preference is a tie in 464/480 (96.6\%). Excluding the rare one-version-acceptable cases (9/480; 1.9\%) changes median \%$\Delta$ by at most 0.3 points and never flips effect directions. We therefore treat the perturbations as semantics-preserving where systematic score shifts reflect evaluator sensitivity rather than perceived mismatch.

\subsection{Evaluation Protocol}
\label{sec:protocol}

All images are scored with \textsc{CLIPS}core, \textsc{PAC-S}, \textsc{UMIC}, \textsc{FLEUR}, and one deterministic judge under a shared preprocessing pipeline. We use paired contrasts wherever applicable (flip vs.\ original; neutral vs.\ modified; TL vs.\ BR) and report:
(i) medians with 95\% BCa bootstrap CIs (10k resamples, seed=2025);
(ii) normality checks via Shapiro-Wilk; paired $t$-test when both samples are approximately normal, else Wilcoxon signed-rank;
(iii) Kruskal-Wallis for multi-level factors (bins, categories, modifier families) with Holm-adjusted pairwise tests;
(iv) effect sizes via Cliff’s $\delta$ for paired deltas.
For comparability to caption evaluators, we also track Spearman/Kendall correlation against \textsc{UMIC}/\textsc{FLEUR}.

\paragraph{Reporting conventions.}
For perturbation analyses, we report the \emph{Median Relative Change},
\[
\%\Delta \;=\; 100 \times \frac{S_{\text{pert}} - S_{\text{orig}}}{S_{\text{orig}}}\,,
\]
aggregated by the median across image-caption pairs, with per-evaluator 95\% BCa bootstrap CIs. Positive $\%\Delta$ means the evaluator scored \emph{higher} under the perturbation; larger $|\%\Delta|$ indicates \emph{worse invariance} (it is not a quality gain).

\subsection{Risk of Ranking Flip (RRF)}
\label{sec:flip}

Small but systematic sensitivities can reorder near-tied systems.
For a fixed evaluator $S$ and a perturbation family $\mathcal{T}$ (e.g., vertical flips), suppose two captioning systems $A$ and $B$ each produce a caption for every image $x$.
Let $S_A(x,t)$ and $S_B(x,t)$ denote the scores that $S$ assigns to the outputs of $A$ and $B$ on image $x$ after applying transform $t\in\mathcal{T}$ (with $t{=}\mathrm{id}$ the unperturbed case), and define the pointwise score gap
\[
\Delta_S(x,t) \;=\; S_A(x,t) - S_B(x,t).
\]
We define the \emph{risk of ranking flip} as

\begin{multline}
\mathrm{RRF}_S(A,B;\mathcal{T})
= \Pr_{x\sim\mathcal{D},\, t\sim\mathcal{T}} \bigl[
\operatorname{sign}\Delta_S(x,t)
\neq {}\\
\operatorname{sign}\Delta_S(x,\mathrm{id})
\bigr],
\end{multline}

i.e., the probability that the ordering between $A$ and $B$ under $S$ changes after applying a semantics-preserving perturbation.
We estimate $\mathrm{RRF}$ using paired bootstraps over images and transforms, and report it as a function of a fixed average score gap $d$ (e.g., $d{=}0.7\%$ on COCO) between $A$ and $B$ on the unperturbed data.
We operationalize near-tied instability via a fixed-gap stress test at $d{=}0.7\%$; full definition and estimation are in Appendix~\ref{app:rq5-extended}.
Intuitively, $\mathrm{RRF}$ answers when given two near-tied systems, how often would their leaderboard order flip under benign changes such as flips or cultural wording?

\subsection{Invariance-Calibrated Scoring}
\label{sec:calibration}


To reduce nuisance sensitivity \emph{without retraining} an evaluator, we post-hoc adjust scores using sensitivities measured under our invariance probes.
Let $S(x,c)$ be the raw score that an evaluator assigns to an image-caption pair $(x,c)$, and let $\mathbb{T}$ denote the set of perturbation families we audit (e.g., spatial, linguistic modifiers).

For a given family $\mathcal{T}\in\mathbb{T}$, we write $(x^{(t)},c^{(t)})$ for the perturbed pair obtained from $(x,c)$ under transform $t\in\mathcal{T}$ (with $t{=}\mathrm{id}$ the original).
We quantify the \emph{sensitivity} of $S$ for $(x,c)$ along $\mathcal{T}$ as
\[
\resizebox{0.98\linewidth}{!}{$
\Delta_S(x,c;\mathcal{T})
=
\operatorname{median}_{t\in\mathcal{T}}
\left|
S\left(x^{(t)},c^{(t)}\right)-S(x,c)
\right|
$}
\]

the median absolute score change across transforms in that family.
The calibrated score is then
\[
\hat{S}(x,c)
\;=\;
S(x,c)
\;-\;
\lambda \sum_{\mathcal{T}\in\mathbb{T}} w_{\mathcal{T}}\,\Delta_S(x,c;\mathcal{T}),
\]
with non-negative weights $w_{\mathcal{T}}$ (default uniform) and a global strength parameter $\lambda\!\geq\!0$.

Intuitively, $\hat{S}$ subtracts nuisance sensitivity estimated from our invariance tests. On a held-out dev set we sweep $\lambda$ and choose the smallest value that substantially reduces median absolute sensitivity while keeping Spearman correlation with learned caption evaluators (UMIC, FLEUR) within 0.01 of the uncalibrated metric. We report both sensitivity reductions and changes in RRF.

\begin{figure}[t]
  \centering
  \includegraphics[width=\linewidth]{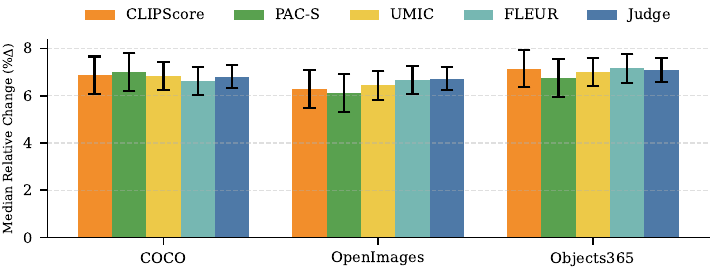}
\caption{\textbf{RQ1a (Vertical flips).} Median \%$\Delta$ (95\% CIs) across evaluators and datasets; positive values indicate higher post-flip scores (orientation 
sensitivity).}

  \label{fig:rq1a}
  \vspace{-1em}
\end{figure}

\section{Experiments}
\label{sec:exp}
We structure our experiments to find answers to targeted questions for the study:

\noindent
\textbf{RQ1 (Spatial invariance)}
How do scores change under flips, repositioning, and light rotations?\\
\textbf{RQ2 (Object sensitivity).}
How do scores vary with object scale and category?\\
\textbf{RQ3 (Socio-linguistic framing).}
How do simple cultural and economic adjectives affect scores relative to neutral controls?\\
\textbf{RQ4 (Cross-dataset behavior).}
Do the trends generalize across metrics and caption suites?\\
\textbf{RQ5 (Ranking impact).}
For near-tied systems, how often do perturbations invert their ranking? 
\textbf{RQ6 (Mitigation).}
Does invariance-calibrated scoring reduce sensitivities and flip risk while preserving correlation with human-aligned caption evaluators?

We apply the perturbation families from Section~\ref{sec:axes} to five evaluators across COCO, OpenImages, and Objects365, replicating key trends on the three caption suites. Each test pairs an unperturbed image-caption with a minimally altered counterpart, so any systematic change reflects metric sensitivity rather than genuine mismatch.

\section{Results}
\label{sec:results}

We investigate patterns and answer RQ1-RQ6 across multiple datasets and metrics. Additional experiments and full details are in Appendix~\ref{app:extended-results}.

\begin{figure}[t]
  \centering
  \includegraphics[width=\linewidth]{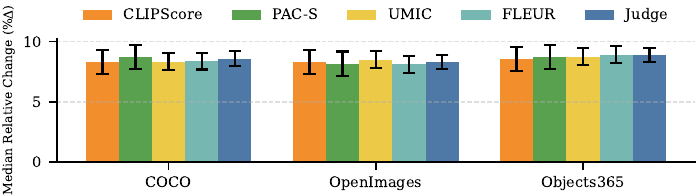}



\caption{\textbf{RQ1b (Repositioning).} Median \%$\Delta$ (BR$-$TL) across evaluators and datasets (95\% CIs). Repositioning induces sizable shifts ($\approx$7–9\%); BR$>$TL.}

  \vspace{-1em}
  \label{fig:rq1b_repos}
\end{figure}

\begin{figure}[t]
  \centering
  \includegraphics[width=\linewidth]{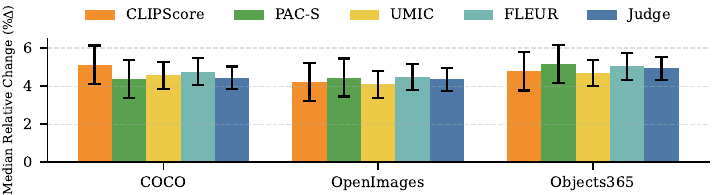}


\caption{\textbf{RQ1b (Rotation).} Median \%$\Delta$ for $\pm10^\circ$ rotations (95\% CIs). Smaller than repositioning ($\approx$4–6\%), yet consistent across evaluators and datasets.}


  \vspace{-1em}
  \label{fig:rq1b_rot}
\end{figure}

\subsection{RQ1: Spatial Invariance}
\label{sec:res-spatial}

\paragraph{Vertical flips (RQ1a).}
Across COCO, OpenImages, and Objects365, all five evaluators increase under vertical flips by $\approx$6–8\% (median \%$\Delta$), with 95\% CIs strictly above zero for every evaluator–dataset pair (Fig.~\ref{fig:rq1a}). Mixed-effects meta-estimates confirm this, with small-to-moderate paired effect sizes (\autoref{tab:flip_summary_v6}). These shifts are large enough to invert near-tied systems (quantified in \S\ref{sec:res-flip}), and the human-validation check (below) rules out perceived semantic change as an explanation.

\paragraph{Context-preserving repositioning and rotations (RQ1b).}
Absolute position in the same scene also matters: moving the dominant object from top-left (TL) to bottom-right (BR) yields the largest spatial deltas ($\approx$7–9\%) across evaluators and datasets (Fig.~\ref{fig:rq1b_repos}), while light in-plane rotations ($\pm10^\circ$) are smaller but reliable ($\approx$4–6\%; Fig.~\ref{fig:rq1b_rot}). Across-quadrant differences are significant (Kruskal–Wallis; Holm-adjusted pairwise tests confirm BR$>$TL; full per-dataset breakdowns in \autoref{tab:rq1b_repos_rot_v6}).

Two checks rule out the explanation that these shifts simply track perceptual quality loss. First, annotators return a majority tie in relative preference on $96.6\%$ of paired items and judge both versions acceptable for $97.3\%$ of items (Appendix~\ref{app:human_validation}); fewer than $2\%$ show a majority-unacceptable version, so spatial perturbations do not make one version systematically worse to humans. Second, a Gaussian-blur control ($\sigma\in\{1.0,2.0\}$) applied to the \emph{original} image reduces detail and naturalness but leaves object position and framing unchanged, and produces markedly smaller shifts than context-preserving repositioning (\autoref{sec:res-object}, \autoref{fig:rq2a}). If naturalness or detail loss were the primary driver, blur would shift scores at least as much as repositioning; it does not. We therefore attribute the observed TL$\to$BR and $\pm 10^{\circ}$ deltas to non-semantic positional sensitivity rather than to human-perceived correctness or naturalness degradation.

\subsection{RQ2: Object Sensitivity}
\label{sec:res-object}

\begin{figure}[!t]
  \centering
  \includegraphics[width=\linewidth]{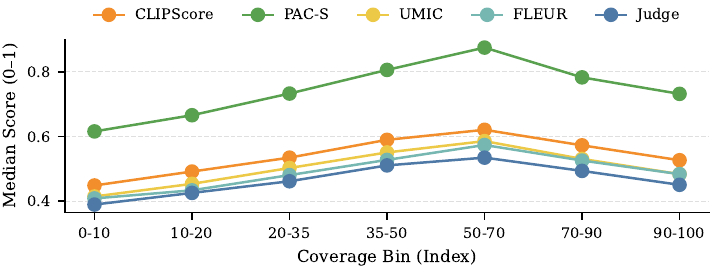}
    \caption{\textbf{RQ2a (Size).} Median score by object coverage bin across evaluators. All metrics peak at 50–70\% coverage and drop at the smallest/largest bins.}

  \vspace{-1em}
  \label{fig:rq2a}
\end{figure}

\paragraph{Scale (RQ2a).}
Across evaluators, scores follow an \emph{inverted-U}: performance peaks at \mbox{50-70\%} coverage and declines at very small/very large targets (Fig.~\ref{fig:rq2a}), with magnitudes larger for embedding similarity than learned caption evaluators. On COCO, extremes are penalized by \mbox{$\approx$6–9\%} relative to the mid-band for embedding metrics and less for learned evaluators; Kruskal–Wallis with Holm-adjusted pairwise tests confirms mid-band $>$ extremes. A blur control on the \emph{original} image yields much smaller shifts, ruling out texture/detail loss as the sole driver.

\begin{figure}[t]
  \centering
  \includegraphics[width=\linewidth]{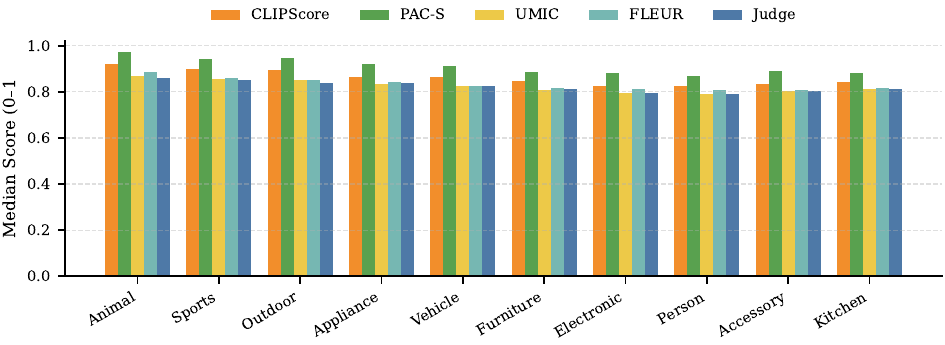}   

  \caption{\textbf{RQ2b (Category).} Median score by category on COCO across evaluators. Category differences are stable across evaluators; \emph{Animal} $>$ \emph{Vehicle} $>$ \emph{Person}.}


  \vspace{-1.5em}
  \label{fig:rq2b}
\end{figure}

\paragraph{Category (RQ2b).}
Under the harmonized taxonomy, macro-median scores differ by several points across categories (Fig.~\ref{fig:rq2b}), with \emph{Animal} near the top and \emph{Person} consistently lower; the spread persists after adjusting for size, implying that category mix can shift headline metrics (Appendix~\ref{app:rq2-extended}).

\subsection{RQ3: Socio-linguistic Framing}
\label{sec:res-soc}

\begin{figure}[t]
  \centering
  \includegraphics[width=\linewidth]{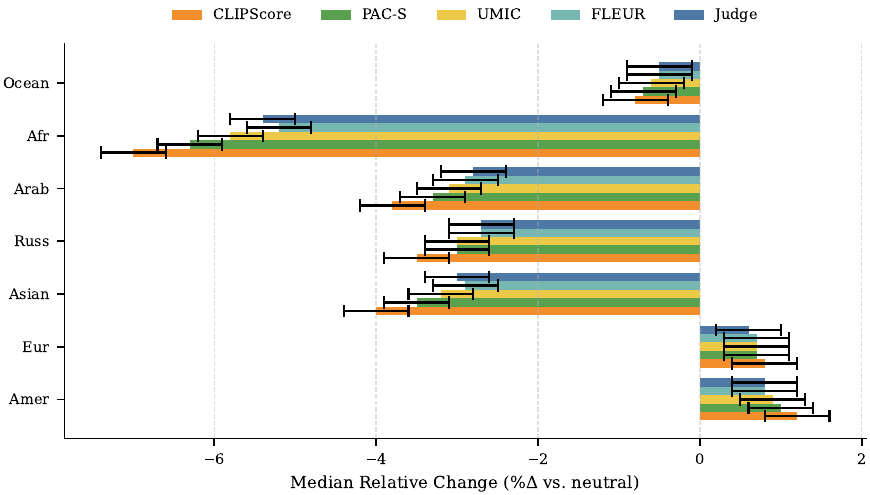}

  \caption{\textbf{RQ3a (Cultural).} Median \%$\Delta$ vs.\ neutral on COCO across evaluators (95\% CIs). Consistent ordering with larger shifts in embedding-similarity metrics.}

  \vspace{-1.5em}
  \label{fig:rq3a}
\end{figure}

\paragraph{Cultural modifiers (RQ3a).}
Relative to neutral phrasing, cultural adjectives induce a consistent ordering with small–to–moderate magnitude on COCO (Fig.~\ref{fig:rq3a}). 
Embedding-similarity evaluators show the strongest effects: for \textsc{CLIPScore}, \emph{African} yields the largest negative median change ($\approx$\,-7\%), while \emph{American}/\emph{European} are mildly positive ($\approx$\,{+}1\%); \textsc{UMIC}/\textsc{FLEUR} and the \textsc{Judge} track the same order at lower magnitude.
Because the image content is unchanged, these shifts reflect sensitivity to non-visual cultural framing rather than scene semantics.
A complementary analysis on naturally occurring captions (Appendix~\ref{app:real_captions}) shows qualitatively similar behavior when we neutralize or swap cultural descriptors in human-written captions: for example, \textsc{CLIPScore} still assigns lower scores on average to captions containing \emph{African} than to matched captions containing \emph{American}/\emph{European}.

\begin{figure}[t]
  \centering
  \includegraphics[width=\linewidth]{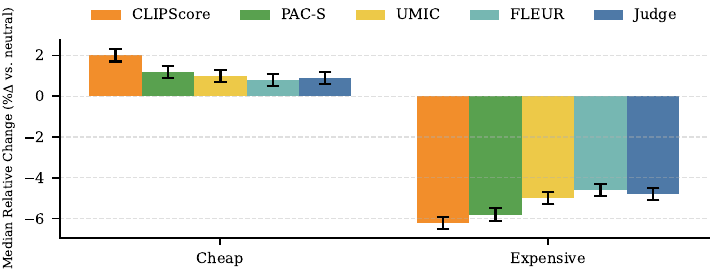}
  
  \caption{\textbf{RQ3b (Economic).} COCO: median \%$\Delta$ vs.\ neutral across evaluators (95\% CIs). \emph{Cheap} is mildly positive, while \emph{expensive} is consistently negative.}

  \vspace{-1em}
  \label{fig:rq3b}
\end{figure}

\paragraph{Economic modifiers (RQ3b).}
For non-person objects, \emph{cheap} is mildly positive while \emph{expensive} is consistently negative across evaluators (Fig.~\ref{fig:rq3b}). 
On COCO, \textsc{CLIPScore} medians are $\approx$\,{+}2.0\% (\emph{cheap}) vs.\ $\approx$\,-6.2\% (\emph{expensive}); \textsc{PAC-S} shows $\approx$\,{+}1.2\% vs.\ $\approx$\,-5.8\%. 
\textsc{UMIC}/\textsc{FLEUR}/\textsc{Judge} display smaller but directionally aligned shifts.
These results indicate that even simple economic framing can systematically alter metric judgments despite identical visuals.
When we apply the same procedure to naturally occurring captions that mention price
(e.g., ``an expensive car'' vs.\ ``a car'' or ``a cheap car''; Appendix~\ref{app:real_captions}),
we again observe that the \emph{median paired score shift} against the neutral caption
follows \emph{cheap} $>$ \emph{neutral} $>$ \emph{expensive} across evaluators.

\paragraph{Embedding-side analysis.}
To understand why CLIP-based metrics exhibit these socio-linguistic asymmetries, we analyze the geometry of the CLIP/PAC-S text encoders (Appendix~\ref{app:text_encoder_bias}).
We construct a simple ``valence'' direction in CLIP text space from clearly positive vs.\ negative adjectives and project each socio-linguistic modifier (and short adjective-noun phrases) onto this direction.
We compute, across modifiers $a$, the Spearman correlation between (i) each modifier’s
projection onto a CLIP text-space ``valence'' direction, $s(a)=\langle e(a),v_{\text{val}}\rangle$,
and (ii) the modifier’s induced median score shift $\widetilde{\Delta}(a)$ (median \%\,$\Delta$
vs.\ the neutral baseline when inserting $a$ into an otherwise unchanged caption).
We obtain $\rho\approx0.7$, meaning modifiers that are more ``negative'' along this direction
tend to produce more negative score shifts, consistent with framing effects being partly driven
by text-encoder geometry.
This suggests that pretraining-induced valence structure in the text encoder 
contributes to the observed effects.

\paragraph{Interpretation.} Our socio-linguistic perturbations use curated adjectives as controlled probes: the image and denoted object are fixed while framing varies along cultural, economic, gender, and affective dimensions. The probes can in principle reveal two distinct failure modes. (a) A \emph{uniform groundedness penalty}, in which the evaluator lowers scores for any adjective it cannot verify against the image; drops would then be symmetric across modifiers of equal groundedness and could be defended as a reasonable evaluation philosophy. (b) \emph{Modifier-identity sensitivity}, in which the evaluator responds asymmetrically to the identity of the modifier, beyond what groundedness alone would predict. What we observe is (b): on the same generic images, where \emph{American bed} and \emph{African bed} are equally ungrounded, \textsc{CLIPScore} yields $\approx +1\%$ for \emph{American}/\emph{European} but $\approx -7\%$ for \emph{African} (\autoref{sec:res-soc}, \autoref{fig:rq3a}). This asymmetry indicates differential priors under controlled equivalence rather than a principled groundedness penalty. The same ordering persists in naturally occurring, and therefore more plausibly grounded, captions when we neutralize or counterfactually swap the modifier (Appendix~\ref{app:real_captions}), and the text-encoder valence analysis (Appendix~\ref{app:text_encoder_bias}) shows these effects are systematically predicted by geometry in the CLIP text space. Annotators do not penalize the modifiers either: both versions are judged acceptable for $>$$97\%$ of items and preferences are majority-tied for $96.6\%$ (Appendix~\ref{app:human_validation}). We read these results as diagnostics of evaluator behavior, not claims about real-world groups; the practical concern is that modifier-identity sensitivity propagates into model selection, ranking, and reward pipelines, where it destabilizes near-tied rankings (\autoref{sec:res-flip}) regardless of whether one frames the sensitivity as a bug or a feature.

\subsection{RQ4: Cross-Dataset behavior}
\label{sec:res-robust}

Patterns from RQ1–RQ3 persist across evaluators and transfer to three caption-evaluation suites (Flickr8k-CF, Pascal-50S, COMPOSITE).
On these external corpora, vertical flips remain \emph{positive} for all evaluator–corpus combinations: medians cluster in the mid–single digits for embedding-similarity metrics and are attenuated yet positive for \textsc{UMIC}/\textsc{FLEUR}/\textsc{Judge}. 
Our judge is instantiated from a GPT-5 model (Appendix~\ref{app:evalcfg}), the fact that it exhibits similar qualitative spatial sensitivities as \textsc{CLIPScore}/\textsc{PAC-S}, albeit at lower magnitude, suggests that these invariance failures are not confined to shallow similarity measures.

\begin{figure}[t]
  \centering
  \includegraphics[width=\linewidth]{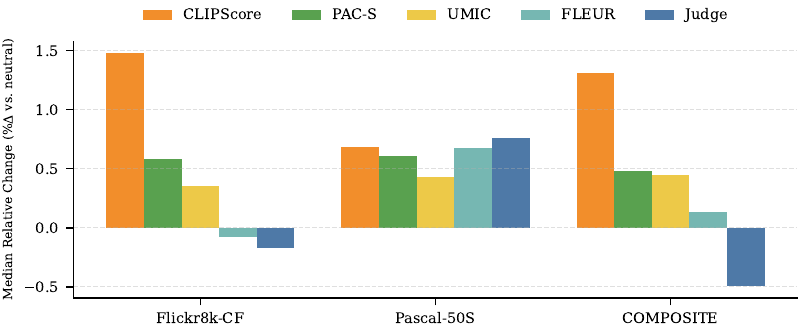}

   \caption{\textbf{RQ4 (External-Cultural).} Results on external corpora (avg.\ \emph{American/European/Asian/African}) highlight persistent effects with reduced magnitude.}





\vspace{-1.5em}
\label{fig:rq4_ext_cult}
\end{figure}

\begin{figure}[t]
  \centering
  \includegraphics[width=\linewidth]{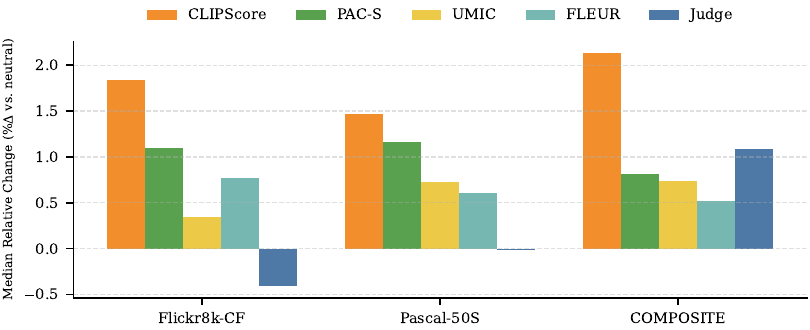}

  \caption{\textbf{RQ4 (External-Economic).} Median \%$\Delta$ vs.\ neutral on external corpora (avg.\ \emph{cheap/expensive}). \emph{Cheap} is mildly positive while \emph{expensive} is negative.
  }
  \vspace{-1em}
  \label{fig:rq4_ext_econ}
\end{figure}

Socio-linguistic framing also replicates: the cultural ordering (\emph{American/European} slightly positive; \emph{African} most negative) and the economic pattern (\emph{cheap} mildly positive; \emph{expensive} negative) hold across corpora (Figs.~\ref{fig:rq4_ext_cult}–\ref{fig:rq4_ext_econ}), again with smaller magnitudes for learned evaluators and the LLM judge.
These results indicate that the sensitivities we document are not an artifact of a single dataset or metric family. Extended results and direction agreement are reported in Appendix~\ref{app:rq4-extended}.

\paragraph{External validity summary.}
Three complementary checks indicate that the sensitivities in RQ1-RQ3 are not artifacts of our templated, single-object regime. \textbf{(i)~Caption-evaluation suites.} RQ1-RQ3 replicate on Flickr8k-CF, Pascal-50S, and COMPOSITE with consistent effect directions across all evaluator--corpus pairs (Figs.~\ref{fig:rq4_ext_cult}--\ref{fig:rq4_ext_econ}; Appendix~\ref{app:rq4-extended}, Tables~\ref{tab:rq4_ext_flips_v8}--\ref{tab:rq4_ext_econ_v8}). \textbf{(ii)~Naturally occurring captions.} On a probe of $732$ human-written captions from MS-COCO and the three suites that already contain our socio-linguistic adjectives (Appendix~\ref{app:real_captions}), neutralization and counterfactual swaps reproduce the same ordering as the templated probes; for example, \textsc{CLIPScore} penalizes captions containing \emph{African} relative to matched \emph{American}/\emph{European} variants by roughly -$5\%$, and \emph{expensive} relative to neutral by -$4\%$ to -$6\%$. \textbf{(iii)~Multi-object pilot.} A $200$-image MS-COCO probe with $2$-$4$ prominent objects (Appendix~\ref{app:multiobject}) shows that spatial effect directions match the single-object slice, with slightly attenuated magnitudes, higher variance, and no qualitative reversals. We therefore treat the audit as a controlled unit-test whose conclusions are supported, though not exhausted, by external-validity evidence on natural captions, multi-object scenes, and standard caption-evaluation suites.

\begin{figure}[t]
  \centering
  \includegraphics[width=\linewidth]{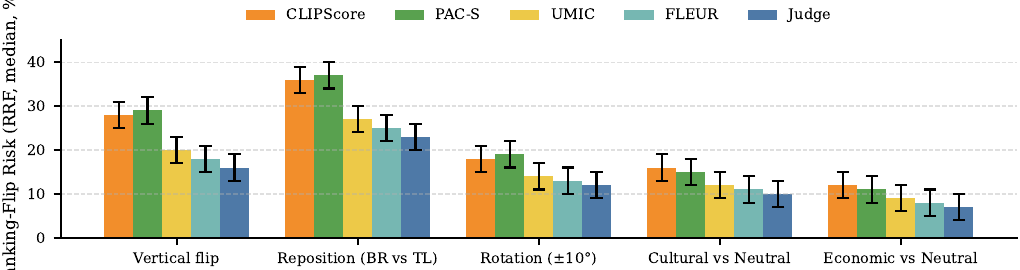}


\caption{\textbf{RQ5 (Flip risk).} Median $\mathrm{RRF}$ (\%) by perturbation family (95\% CIs). Ranking instability is highest under spatial probes, especially repositioning.}

\vspace{-1.2em}
  \label{fig:rq5}
\end{figure}

\subsection{RQ5: Risk of Ranking Flip}
\label{sec:res-flip}

We quantify how often a near-tied leaderboard can invert under semantics-preserving changes using the flip-risk functional $\mathrm{RRF}$ (Section~\ref{sec:flip}), which directly connects to model selection and deployment risk. For a fixed gap $d{=}0.7\%$ on COCO, spatial perturbations (especially \emph{context-preserving repositioning}) produce the largest risks, followed by \emph{cultural} framing; \emph{economic} modifiers and light \emph{rotations} are smaller but clearly non-zero (Fig.~\ref{fig:rq5}, Appendix~\ref{app:rq5-extended}). For \textsc{CLIPScore}, median $\mathrm{RRF}$ is about 28\% (vertical flips), 36\% (repositioning), and 18\% (rotation); socio-linguistic framing is smaller but material ($\approx$16\% cultural, $\approx$12\% economic; Table~\ref{tab:rq5_fliprisk_v6}). \textsc{PAC-S} is similar or slightly higher on spatial axes, while learned evaluators (\textsc{UMIC}/\textsc{FLEUR}) and the judge are consistently lower (e.g., $\approx$23–27\% for repositioning) but still show substantial flip risk. When top systems differ by sub-percent to $\sim$1\%, leaderboard order can invert on roughly one in three evaluation pairs under benign, semantics-preserving changes, most acutely for object repositioning.



\begin{figure}[t]
  \centering
  \begin{minipage}{0.49\linewidth}
    \includegraphics[width=\linewidth]{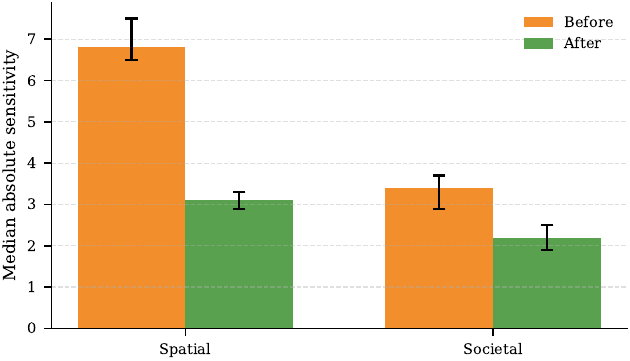}
  \end{minipage}\hfill
  \begin{minipage}{0.49\linewidth}
    \includegraphics[width=\linewidth]{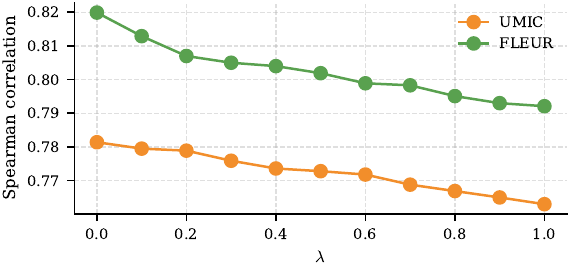}
  \end{minipage}

  \caption{\textbf{RQ6 (Calibration).} Sensitivity by axis before calibration (left) and utility retention via dev-split Spearman vs.\ $\lambda$ (right); $\lambda^\star$ chosen under $\epsilon{=}0.01$ w.r.t.\ \textsc{UMIC}/\textsc{FLEUR}.}
  \vspace{-1em}
  \label{fig:calibration_v6}
\end{figure}

  
  



\subsection{RQ6: Invariance-calibrated scoring}
\label{sec:res-calib}

We apply invariance-calibrated scoring to down-weight non-semantic sensitivity while preserving \emph{agreement} with established caption evaluators.
With a tight agreement constraint ($\epsilon{=}0.01$ vs.\ \textsc{UMIC}/\textsc{FLEUR} on a dev split), calibration reduces \emph{median absolute sensitivity} by roughly half on average across spatial and societal axes; gains are largest on spatial probes (repositioning, flips) and smaller yet consistently non-zero on socio-linguistic probes.
Embedding-similarity evaluators (\textsc{CLIPScore}, \textsc{PAC-S}) benefit most, while \textsc{UMIC}/\textsc{FLEUR} start lower and show smaller reductions.
Figure~\ref{fig:calibration_v6} shows before/after sensitivity.


Calibration also narrows socio-linguistic framing gaps: the median absolute cultural shift across modifiers drops from $\approx$3.4\% to $\approx$2.2\% (35\%), and the spatial gap from $\approx$6.8\% to $\approx$3.1\% (54\%). 
Finally, calibration reduces RRF by lowering the dominant driver: spatial sensitivity, yielding double-digit $\mathrm{RRF}$ drops for repositioning (Table~\ref{tab:calib_rrf_v6}). 
Computationally, calibration is a one-time offline sweep over a fixed set of perturbation variants per dev item; at inference it adds only constant-time post-processing on top of the base metric (Appendix~\ref{app:rq6-extended}). 
To further validate beyond correlations, we evaluate calibrated scores on human-labeled caption-preference suites; calibration preserves (and in some cases slightly improves) pairwise preference accuracy (Appendix~\ref{app:rq6-humanutility}, Table~\ref{tab:rq6_humanutility}). Taken together, invariance-calibrated scoring offers a lightweight mitigation: it reduces nuisance sensitivity and ranking volatility while retaining agreement with standard evaluators and human-preference fidelity. 

The calibration is extensible by design: new invariance axes can be added to $\mathcal{T}$ without retraining the evaluator, with the concrete deployment recipe in Appendix~\ref{app:rq6-extended}. It therefore complements, rather than replaces, ongoing work to broaden the set of audited invariances, and it cannot correct axes that have not yet been instantiated.

\section{Conclusion}
Reference-free captioning evaluators are convenient, but they are not invariant to semantics-preserving changes. Our unit-test framework shows consistent sensitivities across three axes (spatial: orientation, position, light rotation; object: scale, category; socio-linguistic framing: cultural/economic adjectives) and across datasets and metric families, even though a human validation study indicates that these perturbations preserve caption correctness for annotators. These shifts are large enough to destabilize near-tied leaderboards and to induce substantial risk of ranking-flip under benign spatial transforms. We frame these findings as a diagnostic contribution rather than a prescription of a single correct metric objective, whether one treats reference-free evaluators as alignment-focused or typicality-focused, the magnitude and asymmetry of the observed shifts, particularly on socio-linguistic axes, are large enough to warrant explicit disclosure and the kind of post-hoc mitigation we propose.

\noindent\textbf{Practical guidance.}
To make evaluations more robust and comparable, we recommend
(i) reporting \emph{paired} statistics with CIs and effect sizes;
(ii) adding category/scale-stratified summaries;
(iii) using neutral and length-matched controls for wording; and
(iv) adopting simple invariance checks (flip, reposition, rotation) as pre-submission unit tests for metrics and production systems.
Our \emph{invariance-calibrated scoring} provides a post-hoc option that reduces non-semantic sensitivity and RRF while preserving utility correlations.

\noindent\textbf{Outlook.}
Future evaluators should encode spatial awareness, normalize for scale/category composition, and temper language-side framing effects, encouraging bias-aware multimodal evaluation.




\section{Limitations}
\label{sec:limitations}


Our audit is positioned as a controlled unit-test rather than an exhaustive evaluation of reference-free captioning in the wild. The curated single-object slice and templated captions are a deliberate internal-validity choice: by fixing syntax and isolating content to one dominant instance, we can attribute paired score changes to the targeted spatial, object, or socio-linguistic factor rather than to multi-object interactions such as attribute binding, occlusion, co-occurrence, or relational ambiguity. We complement this controlled core with three external-validity checks (replication on three caption-evaluation suites, a multi-object sanity probe, and a socio-linguistic probe on naturally occurring human-written captions), across which the qualitative effects persist. These checks are supporting evidence, not a complete substitute for open-ended evaluation: a full audit on densely compositional scenes, relational captions, and free-form prompting is left to future work.

We study five widely used reference-free evaluators (\textsc{CLIPScore}, \textsc{PAC-S}, \textsc{UMIC}, \textsc{FLEUR}, and one deterministic GPT-5-based judge). While this spans embedding-, learned-, and judge-style scoring, it does not cover the full evaluator space (e.g., alternative CLIP backbones, other learned metrics, or multiple judge prompts/configurations). Our calibration is evaluated on the metrics and probe families in scope; extending it to additional metrics or tasks is a natural next step. The proposed calibration targets invariance-style nuisance factors (e.g., orientation/position and prompt framing) and is not intended to remove object-scale or category effects, which likely reflect systematic evaluator preferences and dataset/model composition.

Our socio-linguistic probes use curated adjective sets as \emph{instruments} to test invariance under controlled wording changes. They are intentionally limited and should not be interpreted as statements about real-world groups or cultures; we report these effects as properties of the evaluators under audit.

Despite these constraints, the findings are consistent across three large sources and persist under neutral and length-matched controls.

\section*{Ethical Considerations}

We use publicly available vision-language datasets (MS-COCO, OpenImages, Objects365, Flickr8k-CF, Pascal-50S, COMPOSITE) under their published licenses and official splits. We release only image identifiers, prompt templates/lexicons, and derived evaluator outputs/analysis artifacts (not the underlying images). Since these corpora may contain sensitive or biased content, our goal is to diagnose evaluator behavior under controlled, semantics-preserving perturbations rather than to validate dataset labels or make claims about depicted subjects.

Our socio-linguistic probes vary caption framing using curated adjective lists (cultural, economic, gender, emotion) while holding the image constant; reported gaps characterize evaluator sensitivity and should not be interpreted as statements about real-world groups. Where human validation is used, it involves low-risk judgments of caption acceptability for synthetic perturbations. We analyze failure modes of widely used reference-free metrics, but we do not propose a single ``correct'' evaluator.
There is a risk that our calibration recipe or flip-risk functional could be misused to optimize leaderboard position without improving reliability or fairness.

\clearpage
\bibliography{custom}

\clearpage
\appendix

\section{Appendix}

\subsection{Extended Related Work}
\label{app:related}

Recent advancements in image captioning evaluation highlight the need to move beyond traditional n‑gram based metrics eg BLEU\cite{papineni2002bleu}, CIDEr \cite{CIDEr}, SPICE\cite{SPICE}, due to their limited semantic fidelity and poor alignment with human judgment. A growing body of research has shifted towards reference‑free metrics that leverage deep multimodal representations and integrate explainability and robustness to perturbations. At the same time, bias and fairness in captioning models and their evaluation have gained significant attention, particularly regarding the amplification of societal biases in both caption generation and evaluation processes.

\subsubsection{Evaluation Metrics for Image Captioning}

Recent advances in image captioning evaluation emphasize moving beyond n-gram metrics such as BLEU\cite{papineni2002bleu}, CIDEr \cite{CIDEr}, and SPICE\cite{SPICE}, which often fall short on semantic fidelity and human alignment. Research has increasingly shifted to reference-free evaluators using pretrained multimodal representations, with added focus on interpretability, robustness to perturbations, and fairness, since both captioning models and metrics can encode and amplify societal biases.

Reference-free metrics have now become central to the evaluation of image captioning models. CLIPScore measures semantic image–text alignment using pretrained CLIP embeddings, offering a reference-free evaluation that correlates strongly with human judgment \citep{hessel2021clipscore}. CLoVe extends the vision-language model framework by refining text-image alignment through object and attribute recognition, making it highly relevant for fine-grained evaluations \citep{castro2024clove}. PickScore fine-tunes CLIP-H on a large dataset of human-generated images and corresponding preferences, improving alignment with user expectations and reflecting human satisfaction \citep{10.5555/3666122.3667716}.

Going beyond scalar scoring, InfoMetIC provides more fine-grained analysis by identifying missing or incorrectly described content, offering better interpretability compared to traditional metrics \citep{hu2023infometic}. ImageReward evaluates the alignment of generated captions with human preferences, directly incorporating human feedback to improve alignment \citep{xu2023imagereward}. Human Preference Score (HPS) fine-tunes the CLIP-L model, focusing on increasing alignment between user-chosen images and textual descriptions, offering a robust metric for subjective quality assessment \citep{wu2023human}.

Robustness to text perturbations has become a key focus. PR-MCS fine-tunes CLIP’s text encoder to maintain stability under lexical perturbations, ensuring better performance across variations in input text \citep{kim2023pr}. Similarly, DENEB reduces sensitivity to hallucinations by training a transformer-based similarity measure on human-annotated data \citep{matsuda2024deneb}. Cobra Effect in Reference-Free Metrics warns that optimizing models against imperfect evaluators can lead to inflated scores that do not genuinely improve the semantic alignment between text and images \citep{ma2024cobra}.

ViLBERTScore evaluates image captioning by using a vision-and-language BERT model to compute textual embeddings for a reference and generated caption. The embeddings are conditioned on the target image, offering a contextual approach to text-image alignment \citep{lee2020vilbertscore}.Complementarily, BRIDGE introduces a learnable multimodal evaluator that better captures visual evidence and text-image alignment by learning from both visual and textual features\cite{sarto2024bridge}.

BLIP-ITC and BLIP2-ITC use contrastive learning for image-text alignment, employing cosine similarity to generate more accurate text-to-image retrieval scores \citep{li2022blip, li2023blip}. TIGEr improves evaluation by integrating text-image grounding, better aligning with human judgment than traditional word-based metrics like BLEU, ROUGE, and METEOR \citep{jiang2019tiger}.  CLIP-R-Precision extends R-Precision by incorporating CLIP embeddings for more human-aligned ranking \citep{park1benchmark, xu2018attngan}. NegCLIP refines text-image alignment by improving CLIP’s ability to reject irrelevant captions \citep{yuksekgonuland}. MosaiCLIP uses scene graphs and graph decomposition to enhance text-image representation, improving alignment by modeling complex relationships between objects \citep{singh2023coarse}.

\subsubsection{Bias, Fairness, and Representational Harm in Captioning}

Bias and fairness are increasingly central to captioning and its evaluation. Understanding and Evaluating Racial Biases in Image Captioning documented demographic disparities in caption quality and content \citep{zhao2021understanding}. Metrics can also be biased: \citep{qiu2023gender} showed CLIPScore may reward gendered language even when inappropriate, reinforcing stereotypes through evaluation. Societal bias amplification work further demonstrates that captioning models can exacerbate biases from training data, motivating careful debiasing strategies \citep{hirota2022quantifying}. ImageCaptioner links these harms to dataset bias (e.g., COCO, Visual Genome) and proposes multimodal protocols to assess bias relative to image content \citep{abdelrahman2024imagecaptioner2}. Broader fairness framing emphasizes representational harms such as misrepresentation and omission \citep{wang2022measuring}, while culturally-aware captioning argues for evaluations that respect diverse contexts and avoid Eurocentric assumptions \citep{yun2024cic}. Mitigation directions include balanced synthetic contrast sets to reduce spurious correlations in evaluation \citep{smith2023balancing}.

Together, these threads motivate evaluation that is both robust to perturbations and sensitive to bias. Our work contributes by introducing a unit-test framework that audits captioning metrics under spatial, object, and linguistic perturbations while probing susceptibility to bias and representational harms in realistic settings.

\subsection{Datasets, Curation, and Balance}
\label{app:balance}

\paragraph{Sources.}
We use MS-COCO 2017 (val), Open Images V7 (val), Objects365 (val), plus three caption-evaluation suites (Flickr8k-CF, Pascal-50S, COMPOSITE) for external validity (Section~\ref{sec:datasets}).

\paragraph{Single-object protocol.}
Images are included when (i) exactly one dominant instance (\textit{max} IoU overlap $<0.1$), (ii) minimal occlusion, (iii) transformations do not truncate the instance, and (iv) normalized area defines one of seven size bins (0–10, 10–20, 20–35, 35–50, 50–70, 70–90, 90–100\%). Rejections and reasons are logged and released.

\begin{figure}[b]
  \centering
  \includegraphics[width=\linewidth]{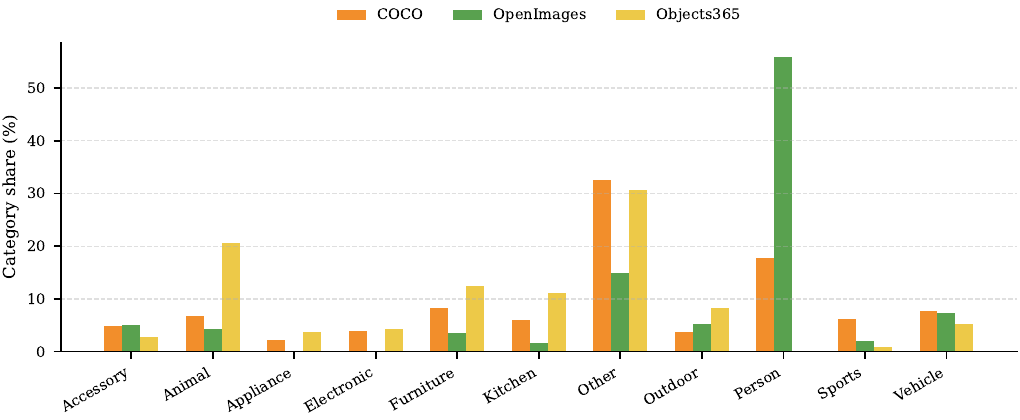}
  \caption{\textbf{Category composition of curated slices.}
  Share (\%) of images per harmonized category in COCO/OpenImages/Objects365.}
  \label{fig:app_taxonomy_dist}
\end{figure}

\paragraph{Taxonomy mapping.}
COCO supercategories, Open Images, and Objects365 labels are mapped into \textit{person}, \textit{animal}, \textit{vehicle}, \textit{furniture}, \textit{kitchen}, \textit{sports}, \textit{electronics}, \textit{indoor}, \textit{outdoor}. Ambiguities are resolved by majority use; ties are flagged and released. Distribution across the taxonomy is depicted in Figure~\ref{fig:app_taxonomy_dist}.


\subsection{Caption Templates and Lexicons}
\label{app:lexicons}

\paragraph{Templates.}
We generate captions from a small set of human-designed templates.
Base captions use a simple declarative form:
\emph{Base}: ``There is a [object].''.
Attribute captions insert a single adjective:
\emph{Attribute}: ``There is a [adjective] [object].''.
Spatial templates add minimal descriptors only when needed for disambiguation (e.g., ``There is a [object] on the left.'') so that spatial wording does not itself encode the downstream perturbations.

\paragraph{Societal adjective families.}
Socio-linguistic adjectives are organized into four primary families plus a small socio-political extension:
(i) \emph{cultural} (e.g., \emph{American}, \emph{European}, \emph{Asian}, \emph{Arab}, \emph{African}, \emph{Russian}, \emph{Oceanian});
(ii) \emph{economic} (e.g., \emph{cheap}, \emph{expensive}, \emph{luxury}, \emph{budget});
(iii) \emph{gender} (e.g., \emph{male}, \emph{female}, \emph{boy}, \emph{girl});
(iv) \emph{emotion} (e.g., \emph{happy}, \emph{sad}, \emph{angry}); and
(v) a small set of \emph{occupational / socio-political} descriptors (e.g., \emph{local}, \emph{foreign}, \emph{immigrant}, \emph{citizen}, \emph{refugee}, \emph{tourist}) used only in targeted probes.
Each family includes neutral and length-matched controls such as \emph{typical}, \emph{plain}, or \emph{ordinary}, chosen to match the number of characters and tokens as closely as possible.

We construct adjective–noun pairs by combining these modifiers with compatible object types.
For a subset of analyses we build simple intersectional combinations (e.g., \emph{African man}, \emph{American car}, \emph{expensive sofa}) to test whether sensitivities differ between person vs.\ non-person categories and between object types.

\paragraph{Compatibility screen.}
We filter ill-formed adjective–noun pairs (e.g., POS conflicts, animacy mismatches, or implausible combinations such as \emph{angry chair}) using a set of heuristics and manual spot checks.
The compact main subset used in the body is released alongside the \emph{rejected} pairs and their rejection reasons.


\subsection{Perturbation Implementations}
\label{app:perturb_impl}

Table~\ref{tab:perturbations} summarizes the perturbation families and the invariance properties they probe and the implementation details are as follows -

\begin{table}[t]
\centering

\setlength{\tabcolsep}{4pt}
\renewcommand{\arraystretch}{1.1}
\begin{tabularx}{\columnwidth}{p{1.3cm}YY}
\toprule
\textbf{Axis} & \textbf{Perturbations} & \textbf{Probes} \\
\midrule
Spatial  & Flips; context-preserving repositioning; $\pm10^{\circ}$ rotations; distance-to-center; blur control & Invariance to orientation/layout; texture/detail sensitivity \\
Object   & Seven coverage bins; harmonized categories & Scale effects; category composition \\
Societal & Cultural/economic/ gender/emotion with neutral and length-matched controls & Sensitivity to non-visual framing \\
\bottomrule
\end{tabularx}
\caption{Perturbations and targeted properties.}
\label{tab:perturbations}
\end{table}

\paragraph{Flips and rotations.}
Vertical/horizontal flips are image-space transforms; rotations use $\pm 5^\circ,\pm 10^\circ$ with reflection padding and center-preserving resampling.

\paragraph{Context-preserving repositioning.}
We preserve background and scale, translating the segmented dominant object; collisions/occlusions trigger resampling until a valid placement is found (Algorithm~\ref{alg:anchor_reloc}).

\begin{algorithm*}[t]
\caption{Anchor-Relocation Composition}
\label{alg:anchor_reloc}
\begin{algorithmic}[1]
\Require image $I$, mask $M$ (dominant object), anchors $A=\{\text{TL},\text{TR},\text{BL},\text{BR}\}$
\State $P \gets I \odot M$ \Comment{object patch}
\State $B \gets I \odot (1-M)$ \Comment{background}
\For{$a \in A$}
  \State $t \gets \textsc{TranslateCentroidToAnchor}(P,a)$
  \While{$t$ collides with image boundary \textbf{or} occludes salient regions}
    \State $t \gets \textsc{ResampleJitterTowardCenter}()$ \Comment{up to $K$ tries}
  \EndWhile
  \State $I_a \gets B \oplus \textsc{Translate}(P,t)$ \Comment{feathered alpha to avoid hard edges}

\EndFor
\State \Return $\{I_a\}_a$
\end{algorithmic}
\end{algorithm*}

\paragraph{Repositioning artifact diagnostics.}
We compute (i) a background-change score as the mean absolute pixel difference outside the union of the source and target object regions, and (ii) a boundary-seam score using edge energy in a thin ring around the pasted mask boundary.
We re-run RQ1b after removing the top 5\% (and 10\%) highest-seam samples; effect directions and medians remain stable. 

\paragraph{Distance-to-center proxy.}
We use coverage bins as a coarse proxy (main text, Fig.~\ref{fig:rq2a}); a continuous sweep with cubic splines is provided in the released notebooks.

\subsection{Extended Experimental Setup}
\label{app:evalcfg}

\paragraph{Shared preprocessing.}
All evaluators use the same resized-crop resolution and normalization; captions pass through the same tokenizer. We log image IDs, seeds, and exact scripts.

\paragraph{CLIPScore / PAC-S.}
\textsc{CLIPScore} computes the cosine of pooled image/text embeddings; \textsc{PAC-S} uses CLIP feature similarity with perceptual calibration. Backbones and pooling choices are recorded.

\paragraph{UMIC / FLEUR.}
We use publicly released caption-quality evaluators; we record model version/commit and interface limits.

\paragraph{Judge.}
Our judge is instantiated from a GPT-5 model exposed through an evaluation API.
We use a frozen rubric and system prompt that instructs the model to score how well a caption describes an image on a 0–10 scale, with emphasis on factual correctness and relevance.
Inference is deterministic (temperature $T{=}0$, fixed seed, no sampling from multiple candidates), and runs are cached by (imageID, caption, rubric) so that repeated uses are bitwise identical.
We treat this judge as representative of mainstream LLM-based evaluation practice and report it alongside \textsc{CLIPScore}, \textsc{PAC-S}, \textsc{UMIC}, and \textsc{FLEUR} throughout.

\paragraph{Implementation details.}
All evaluators run deterministically on identical pre-processing; the judge uses a fixed rubric, temperature $=0$, and a seeded context. Bootstrap uses $10{,}000$ resamples (seed $=2025$).

\subsection{Statistical Procedures and Meta-Analysis}
\label{app:stats}

\paragraph{Paired contrasts.}
Medians of paired differences with 95\% BCa bootstrap CIs (10k resamples, fixed seed). Normality via Shapiro–Wilk; two-tailed paired $t$-test when applicable, otherwise Wilcoxon.

\paragraph{Multi-level factors and multiplicity.}
Kruskal–Wallis across bins/categories; Holm-adjusted pairwise tests.

\paragraph{Effect sizes and mixed-effects.}
Cliff’s $\delta$ for paired deltas. Random-intercept mixed-effects models report $\beta_1$ (95\% CI) alongside nonparametrics.

\subsection{Calibration Ablations and Alternatives}
\label{app:calib}

\paragraph{Selecting $\lambda^\star$ under a correlation constraint.}
We choose $\lambda^\star$ on a development split by grid search
$\lambda\in\{0,0.05,\ldots,1.0\}$, minimizing total non-semantic sensitivity
$\sum_{\mathcal{T}\in\mathbb{T}} \operatorname{median}\!\left|\Delta_{\hat{S}}(\cdot;\mathcal{T})\right|$
subject to correlation preservation:
$\mathrm{corr}(\hat{S},H)\ge \mathrm{corr}(S,H)-\epsilon$ for $H\in\{\textsc{UMIC},\textsc{FLEUR}\}$.
The main text (Fig.~\ref{fig:calibration_v6}) reports (i) aggregate before/after median absolute sensitivity by axis and
(ii) correlation vs.\ $\lambda$ with the selected $\lambda^\star$ markers.

\paragraph{Per-metric outcomes.}
We report for each evaluator: (a) axis-wise median absolute sensitivity (spatial/object/societal) before$\to$after,
(b) the selected $\lambda^\star$, and (c) the corresponding $\Delta$correlation with \textsc{UMIC}/\textsc{FLEUR}.

\paragraph{Ablations.}
We verify that the calibration behavior is not sensitive to minor design choices:
\begin{itemize}\setlength\itemsep{2pt}
  \item \textbf{Axis weights.} Uniform weights vs.\ sensitivity-proportional weights
  ($w_{\mathcal{T}}\propto$ baseline $\operatorname{median}|\Delta|$) yield similar reductions at comparable $\Delta$correlation.%
  (see \texttt{tables/calib\_axisweights.csv}).
  \item \textbf{Constraint strength.} We sweep $\epsilon\in\{0.005,0.01,0.02\}$; chosen $\lambda^\star$ shifts as expected
  but sensitivity reductions remain qualitatively stable.
  \item \textbf{Alternative formulations.} We test a ridge-style shrinkage and a quantile-subtraction variant;
  both reduce sensitivity but are slightly less stable than the default across evaluators.
\end{itemize}





\subsection{Repositioning transform diagnostics}
\label{app:repositioning_diagnostics}

\paragraph{Motivation.}
Context-preserving repositioning composes a segmented object patch at a new location on the \emph{same} background.
Because imperfect masks or blending could introduce low-level artifacts, we run diagnostics to verify that (i) the background is unchanged away from the moved object and (ii) our RQ1b effects persist after removing artifact-heavy composites.

\paragraph{Background preservation (BG$\Delta$).}
Let $I$ be the original image and $I'$ the repositioned image (RGB normalized to $[0,1]$).
Let $M_s$ be the source object mask in $I$ and $M_t$ the target mask location in $I'$.
We exclude a small neighborhood around either object region and compute mean absolute background change:
\[
\begin{aligned}
U &= \mathrm{dilate}(M_s \cup M_t, r_{\text{bg}}),\\
\mathrm{BG}\Delta(I,I') &=
\frac{1}{|U^c|}\sum_{p\in U^c}\left\|I(p)-I'(p)\right\|_1 ,
\end{aligned}
\]

where $\|\cdot\|_1$ sums absolute differences across channels.
Low BG$\Delta$ indicates background preservation up to negligible interpolation.

\paragraph{Boundary seam strength (SeamE).}
To quantify potential halo/seam artifacts at the pasted boundary, we measure Sobel gradient energy in a thin ring around the target mask:

\[
\begin{aligned}
R &= \mathrm{dilate}(M_t,r_2)\setminus \mathrm{dilate}(M_t,r_1),\\
\mathrm{SeamE}(I') &=
\frac{1}{|R|}\sum_{p\in R} G(I'(p)),
\end{aligned}
\]

where $G(\cdot)$ is gradient magnitude (computed per-channel and averaged).
We report \emph{relative} seam strength as a ratio $\mathrm{SeamE}(I')/\mathrm{SeamE}(I)$, where $\mathrm{SeamE}(I)$ is computed on an analogous ring around $M_s$ in the original image; ratios near 1 indicate no systematic seam amplification beyond natural object boundaries.

\paragraph{Robustness to artifact filtering and blending.}
We recompute the RQ1b repositioning effect (median \%$\Delta$ for BR$-$TL with the same BCa protocol as main experiments) after filtering the top $q\%$ highest-scoring samples by BG$\Delta$, by SeamE, or by either.
We also re-render a stratified subset using a \emph{feathered alpha} boundary (a 3px linear ramp) to soften hard cut edges.
Across filters and blending choices, effect directions are unchanged and medians remain within overlapping confidence intervals, indicating that the repositioning sensitivity is not driven by a small set of artifact-heavy composites.


\begin{table*}[t]
  \centering
  \scriptsize
  \setlength{\tabcolsep}{3pt}
  \begin{tabular}{lccccccccc}
    \toprule
    \textbf{Condition} & \textbf{Ret.} &
    \textbf{BG$\Delta$ ($\times10^3$)} &
    \textbf{SeamE ratio} &
    \textbf{CLIPScore} &
    \textbf{PAC-S} &
    \textbf{UMIC} &
    \textbf{FLEUR} &
    \textbf{Judge} \\
    \midrule
    None (all)                 & 100\% & 0.25 & 1.01 & 8.4 & 8.7 & 7.6 & 7.2 & 6.6 \\
    Drop top 5\% SeamE         &  95\% & 0.24 & 0.99 & 8.3 & 8.6 & 7.5 & 7.1 & 6.5 \\
    Drop top 5\% BG$\Delta$    &  95\% & 0.18 & 1.01 & 8.4 & 8.7 & 7.6 & 7.2 & 6.6 \\
    Drop top 5\% either        &  92\% & 0.18 & 0.99 & 8.2 & 8.5 & 7.4 & 7.0 & 6.4 \\
    Drop top 10\% either       &  85\% & 0.15 & 0.98 & 8.1 & 8.4 & 7.3 & 6.9 & 6.3 \\
    Feathered-alpha subset     &  20\% & 0.26 & 0.96 & 8.3 & 8.6 & 7.5 & 7.1 & 6.5 \\
    \bottomrule
  \end{tabular}
  \caption{\textbf{Diagnostics for context-preserving repositioning (RQ1b).}
  BG$\Delta$ is mean absolute background change outside a dilated union of source/target masks (lower is better; values shown $\times10^3$ for readability).
  SeamE ratio compares boundary edge energy in $I'$ to the analogous ring in $I$ (ratio $\approx 1$ indicates no seam amplification).
  Right columns report the resulting median \%$\Delta$ (BR$-$TL) per evaluator after filtering/blending; values remain stable, supporting that RQ1b is not driven by compositing artifacts.}
  \label{tab:repositioning_diagnostics}
\end{table*}

\paragraph{Fixed parameters.}
We use $r_{\text{bg}}=8$ px, $(r_1,r_2)=(2,5)$ px for the seam ring, and $q\in\{5,10\}$ for filtering, fixed \emph{a priori} and applied uniformly across datasets/evaluators.
``Feathered alpha'' uses a 3px linear ramp at the mask boundary to reduce hard-edge transitions.

\subsection{External Validity on Caption Evaluation Suites}
\label{app:external}

We replicate RQ1-RQ3 on Flickr8k-CF, Pascal-50S, and COMPOSITE using their image pools and pairings. To limit domain shift, we apply spatial flips and a reduced cultural/economic subset. Effect directions replicate; magnitudes vary with caption style. 

\subsection{Multi-object sanity check}
\label{app:multiobject}

\paragraph{Setup.}
To partially address the gap between our single-object regime and real-world multi-object scenes, we run a small sanity check on a COCO subset with two to four prominent objects.
We sample 200 images that pass basic quality filters (no extreme occlusions, reasonably sized instances) and construct captions using the same templated scheme (e.g., ``There is a [object].'') targeting the dominant object.
We then apply the same spatial perturbations as in RQ1 (vertical flips, context-preserving repositioning, light rotations) to the full image while keeping the caption fixed.

\paragraph{Results.}
Across evaluators, the direction of effects matches the single-object slice: vertical flips and repositioning systematically increase scores, with magnitudes slightly attenuated but within the confidence bands of the single-object results; rotations show smaller but consistent positive shifts.
Because the multi-object images introduce additional clutter and interacting objects, variance is higher and some per-category effects become noisier, but we do not observe any reversals in the qualitative patterns.

\paragraph{Interpretation.}
This probe is deliberately small and not a full multi-object audit, but it suggests that the spatial sensitivities documented in the main text are not an artifact of the single-object regime.
A more thorough treatment of cluttered scenes and relational captions is left for future work.



\subsection{Human validation of perturbations}
\label{app:human_validation}

\paragraph{Sampling.}
We annotate $N_\text{human}=480$ paired items from the curated COCO single-object slice, stratified by perturbation family:
(1) \textbf{Spatial} ($n=160$): 80 vertical flips, 40 TL$\leftrightarrow$BR repositionings, and 40 $\pm10^\circ$ rotations.
(2) \textbf{Object/texture control} ($n=160$): Gaussian-blur controls applied to images drawn from size extremes (80 smallest-bin, 80 largest-bin);
(3) \textbf{Socio-linguistic} ($n=160$): 40 each for cultural, economic, gender, and emotion modifiers.
Each item contains two versions of the same underlying example. For spatial/object items we pair the original and perturbed \emph{image} with a shared caption; for socio-linguistic items we use a shared image with a neutral vs.\ modified caption. The two versions are shown in randomized left/right order.

\paragraph{Annotation protocol.}
Annotators (English-speaking; 3 per item) answer:
(i) \textbf{Acceptability (per version).} ``How well does this caption describe the image?'' with options
\emph{incorrect}, \emph{partially correct}, \emph{fully correct}. We map \emph{partially/fully correct} to \emph{acceptable} and \emph{incorrect} to \emph{unacceptable}.
(ii) \textbf{Relative preference.} ``Which is better, or are they equally good?'' with options \emph{A}, \emph{B}, \emph{Tie}.

\paragraph{Aggregate outcomes.}
Using majority vote ($\ge2/3$), 467/480 items (97.3\%) are \emph{acceptable for both} versions; 9/480 (1.9\%) are acceptable for only one version; the remainder have no majority or both unacceptable. For relative preference, 464/480 items (96.6\%) are majority \emph{tie}; non-tie preferences are roughly balanced between the two versions. Inter-annotator agreement is substantial (Fleiss' $\kappa=0.63$ for acceptability; $\kappa=0.59$ for preference).


\paragraph{Effect on reported sensitivities.}
We perform two complementary robustness checks on the aggregate
outcomes. First, we recompute the main spatial and socio-linguistic
analyses after excluding the $9$ items where only one version is
majority-acceptable; across evaluator--dataset pairs, median
\%$\Delta$ changes by at most $0.3$ percentage points and all effect
directions are preserved. Second, as a stricter check, we also
remove the $3.1\%$ ($15/480$) of items for which either version was
marked only \emph{partially correct} by the majority of annotators,
so that the remaining set contains items judged \emph{fully correct}
on both versions. Recomputing the main analyses on this stricter
subset leaves medians, effect directions, and the outcomes of the
significance tests (paired Wilcoxon / Kruskal--Wallis with
Holm adjustment) unchanged up to the same $\leq 0.3$ percentage
point tolerance. Together, these two checks indicate that the
reported metric sensitivities are not driven by a small set of
non-equivalent pairs or by items near the acceptability boundary,
and support treating our perturbations as semantics-preserving for
humans.

\subsection{Socio-linguistic framing probe}
\label{app:real_captions}

\paragraph{Data selection.}
To test whether the socio-linguistic sensitivities observed with templated captions also appear for naturally occurring language, we search MS-COCO and the three caption-evaluation suites (Flickr8k-CF, Pascal-50S, COMPOSITE) for captions that contain adjectives from our socio-linguistic lexicons (Appendix~\ref{app:lexicons}).
We focus on cultural (e.g., \emph{American}, \emph{African}), economic (e.g., \emph{cheap}, \emph{expensive}), gender, and emotion modifiers.
After filtering for clear, single-object descriptions and removing duplicates, we obtain 732 base captions across corpora (412 cultural, 196 economic, 124 gender/emotion).

\paragraph{Neutralization and counterfactual rewrites.}
For each base caption $c_\text{orig}$ containing a socio-linguistic adjective $a$ modifying an object $o$, we construct:

\begin{itemize}
    \item a \textbf{neutralized} caption $c_\text{neu}$ by replacing $a$ with a neutral, length-matched control (e.g., \emph{typical}, \emph{plain});
    \item an \textbf{alternate} caption $c_\text{alt}$ where we swap $a$ for an antonym or contrasting modifier from the same family (e.g., \emph{African}$\rightarrow$\emph{American}/\emph{European}; \emph{cheap}$\rightarrow$\emph{expensive}).
\end{itemize}

We keep the remainder of the caption identical and reuse the original image.
We discard rewrites that become ungrammatical or pragmatically implausible under manual spot checks.

\paragraph{Results.}
We run the same five evaluators on $(x, c_\text{orig})$, $(x, c_\text{neu})$, and $(x, c_\text{alt})$.
For cultural adjectives, \textsc{CLIPScore} and \textsc{PAC-S} show the same ordering as in the templated setting: captions containing \emph{African} are scored lower on average than matched captions containing \emph{American}/\emph{European}, with median relative changes of roughly -5\% vs.\ neutralized variants; \emph{American}/\emph{European} are mildly positive ($\approx$+0.5–1.0\%) relative to neutral.
\textsc{UMIC} and \textsc{FLEUR} attenuate magnitudes but preserve the ordering.
For economic descriptors, \emph{cheap} captions are again mildly positive (median $\approx$+1–2\%) and \emph{expensive} captions negative (median $\approx$-4–6\%) relative to their neutralized counterparts, with learned evaluators closer to zero but directionally aligned.

These magnitudes are smaller than those observed under templated captions (Section~\ref{sec:res-soc}, Appendix~\ref{app:societal_appendix}), reflecting the greater linguistic and contextual variability of real captions, but the qualitative patterns persist.
This complementary probe suggests that the socio-linguistic framing sensitivities we document are not an artifact of our templating scheme and can arise in naturally written descriptions as well.

\subsection{Text-encoder embedding analysis}
\label{app:text_encoder_bias}

To probe whether the socio-linguistic sensitivities documented in Section~\ref{sec:res-soc} are reflected in the underlying text encoders, we analyze CLIP/PAC-S embeddings for the adjectives used in our cultural, economic, gender, and emotion families.

\paragraph{Setup.}
We use the same CLIP text backbone used by \textsc{CLIPScore} and \textsc{PAC-S}.
For each adjective $a$ (e.g., \emph{African}, \emph{American}, \emph{cheap}, \emph{expensive}), we compute its normalized text embedding $e(a)$ as well as phrase embeddings $e(a,o)$ for simple adjective–noun phrases (e.g., ``African bed'', ``expensive car'') where $o$ ranges over a representative subset of objects from our taxonomy.
We also construct a simple ``valence'' direction in text space,
\[
\begin{aligned}
v_{\mathrm{val}} &=
\tfrac{1}{n}\big(e(\text{``good''}) + e(\text{``nice''})\\&\quad + ... + e(\text{``beautiful''})\big) \\
&\quad - \tfrac{1}{n}\big(e(\text{``bad''}) \\&\quad + ... +  e(\text{``ugly''}) + e(\text{``terrible''})\big),
\end{aligned}
\]


and normalize $v_{\mathrm{val}}$ to unit length.
For each adjective (or phrase), we compute the scalar projection
$s(a) = \langle e(a), v_{\mathrm{val}} \rangle$ (or $s(a,o) = \langle e(a,o), v_{\mathrm{val}} \rangle$).

\paragraph{Cultural and economic adjectives.}
We then correlate these projections with the median relative change \%$\Delta$ induced by the corresponding modifiers when inserted into captions (Tables~\ref{tab:rq3a_cultural_COCO_v6}-\ref{tab:rq3b_economic_COCO_v6}).
Across the seven cultural adjectives on COCO (American, European, Asian, Arab, African, Russian, Oceanian), we observe a strong monotone relationship between $s(a)$ and the median \%$\Delta$ for \textsc{CLIPScore}: Spearman $\rho\!\approx\!0.70$ ($p<10^{-2}$) at the word level and $\rho\!\approx\!0.78$ for phrase embeddings $e(a,o)$ averaged over objects $o$.
\textsc{PAC-S} exhibits a similar pattern with $\rho\!\approx\!0.66$ (word) and $\rho\!\approx\!0.73$ (phrase).
Adjectives with more negative projections on $v_{\mathrm{val}}$ (e.g., \emph{African}, \emph{cheap}, negative emotions) systematically yield lower metric scores when used in captions, whereas adjectives with more positive projections (\emph{American}, \emph{European}, \emph{happy}) are associated with higher scores.

\begin{table*}[!t]
\centering
\footnotesize
\begin{tabular}{lcccc}
\toprule
\textbf{Evaluator} & \textbf{COCO} & \textbf{OpenImages} & \textbf{Objects365} & \makecell{\textbf{Meta} \\ $\beta_1$ [CI]} \\
\midrule
CLIPScore & 6.9\% [6.1,7.7]~($\delta$ 0.28) & 6.3\% [5.5,7.1]~($\delta$ 0.28) & 7.2\% [6.4,8.0]~($\delta$ 0.28) & 0.069 t[0.061,0.077] \\
PAC-S & 7.0\% [6.2,7.8]~($\delta$ 0.28) & 6.1\% [5.3,6.9]~($\delta$ 0.28) & 6.7\% [5.9,7.5]~($\delta$ 0.28) & 0.070 [0.062,0.078] \\
UMIC & 6.8\% [6.2,7.4]~($\delta$ 0.20) & 6.4\% [5.8,7.0]~($\delta$ 0.20) & 7.0\% [6.4,7.6]~($\delta$ 0.20) & 0.068 [0.062,0.074] \\
FLEUR & 6.6\% [6.0,7.2]~($\delta$ 0.20) & 6.7\% [6.1,7.3]~($\delta$ 0.20) & 7.2\% [6.6,7.8]~($\delta$ 0.20) & 0.066 [0.060,0.072] \\
Judge & 6.8\% [6.3,7.3]~($\delta$ 0.16) & 6.7\% [6.2,7.2]~($\delta$ 0.16) & 7.1\% [6.6,7.6]~($\delta$ 0.16) & 0.068 [0.063,0.073] \\
\bottomrule
\end{tabular}
\caption{RQ1a: Vertical flip sensitivity (\%$\Delta$). Median [95\% CI]; Cliff's $\delta$; mixed-effects $\beta_1$ (REML).}
\label{tab:flip_summary_v6}
\vspace{-1.5em}
\end{table*}

For economic modifiers, \emph{cheap} and \emph{expensive} cluster on opposite sides of the valence direction: \emph{cheap} lies slightly above the neutral baseline (\emph{typical}/\emph{plain}), while \emph{expensive} lies below.
This matches the metric behavior in Table~\ref{tab:rq3b_economic_COCO_v6} and Figure~\ref{fig:rq3b}: \emph{cheap} is mildly positive relative to neutral, whereas \emph{expensive} is strongly negative.

\paragraph{Gender and emotion.}
Applying the same analysis to gender (\emph{male}/\emph{female}) and emotion (\emph{happy}/\emph{sad}/\emph{angry}) adjectives yields analogous results.
For emotion, the projection scores order as \emph{happy} $>$ neutral $>$ \emph{sad}/\emph{angry}, and this ordering matches the median \%$\Delta$ in Figure~\ref{fig:rq3_emotion}.
Correlations are again higher for phrase-level embeddings (e.g., ``sad person'') than for adjectives alone.

\paragraph{Interpretation.}
These findings suggest that CLIP’s text encoder encodes socio-linguistic descriptors along a latent valence axis in a way that is predictive of how \textsc{CLIPScore}/\textsc{PAC-S} respond to them in caption scoring.
In other words, the socio-linguistic framing effects documented in Section~\ref{sec:res-soc} are not arbitrary artifacts of our templating scheme but emerge from systematic structure in the underlying text representation.
This supports the view that pretraining-induced priors over cultural, economic, and affective descriptors can directly influence downstream evaluation metrics built on these encoders.

\section{Extended Results}
\label{app:extended-results}

This appendix provides extended results for all research questions (RQ1–RQ5), including additional probes and per-dataset breakdowns. We document the exact statistical tests (with corrections) and include supporting evidence to substantiate the claims in the main text.

\subsection{RQ1: Spatial Invariance}
\label{app:rq1-extended}

\paragraph{Statistical Procedure.}
All contrasts are paired at the image level. We report medians of paired differences with 95\% BCa bootstrap CIs (10{,}000 resamples, fixed seed). Normality is screened with Shapiro–Wilk; when non-normality is detected we use Wilcoxon signed-rank (two-sided), otherwise paired $t$.
Across multi-level factors (quadrants), we use Kruskal–Wallis with Holm-adjusted pairwise comparisons.
Mixed-effects models use random intercepts for \emph{image} and \emph{category} (REML) and report the perturbation coefficient $\beta_1$ with 95\% CIs; these complement the nonparametric summaries.

\paragraph{Vertical flips (RQ1a).}
Per-dataset medians, 95\% CIs, Cliff’s $\delta$, and mixed-effects meta-estimates appear in \autoref{tab:flip_summary_v6}.
Across all evaluator–dataset cells, medians are positive with CIs excluding zero, indicating a consistent increase after a vertical flip.
Effect sizes are small-to-moderate but practically meaningful given RQ5 (RRF).

\begin{table}[!ht]
\centering
\footnotesize

\resizebox{\columnwidth}{!}{%
\begin{tabular}{@{}lcccc@{}}
\toprule
\textbf{Evaluator} & \textbf{COCO} & \textbf{OpenImages} & \textbf{Objects365} & \textbf{Type} \\
\midrule
CLIPScore & 8.3\% [7.3,9.3] & 8.3\% [7.3,9.3] & 8.6\% [7.6,9.6] & Reposition \\
CLIPScore & 5.1\% [4.1,6.1] & 4.2\% [3.2,5.2] & 4.8\% [3.8,5.8] & Rotation \\
PAC-S & 8.7\% [7.7,9.7] & 8.2\% [7.2,9.2] & 8.8\% [7.8,9.8] & Reposition \\
PAC-S & 4.4\% [3.4,5.4] & 4.5\% [3.5,5.5] & 5.2\% [4.2,6.2] & Rotation \\
UMIC & 8.3\% [7.6,9.0] & 8.5\% [7.8,9.2] & 8.8\% [8.1,9.5] & Reposition \\
UMIC & 4.6\% [3.9,5.3] & 4.1\% [3.4,4.8] & 4.7\% [4.0,5.4] & Rotation \\
FLEUR & 8.4\% [7.7,9.1] & 8.1\% [7.4,8.8] & 8.9\% [8.2,9.6] & Reposition \\
FLEUR & 4.8\% [4.1,5.5] & 4.5\% [3.8,5.2] & 5.0\% [4.3,5.7] & Rotation \\
Judge & 8.6\% [8.0,9.2] & 8.3\% [7.7,8.9] & 8.9\% [8.3,9.5] & Reposition \\
Judge & 4.4\% [3.8,5.0] & 4.4\% [3.8,5.0] & 4.9\% [4.3,5.5] & Rotation \\
\bottomrule
\end{tabular}}
\caption{RQ1b: Context-preserving repositioning and light rotations (\%$\Delta$). Median [95\% CI].}
\label{tab:rq1b_repos_rot_v6}
\vspace{-1.5em}
\end{table}

\paragraph{Context-preserving repositioning and rotations (RQ1b).}
\autoref{tab:rq1b_repos_rot_v6} reports per-dataset medians and 95\% CIs for (i) context-preserving relocation from top-left (TL) to bottom-right (BR) at constant scale, and (ii) light in-plane rotations ($\pm10^\circ$).
Across quadrants, Kruskal–Wallis detects differences for every evaluator; Holm-adjusted pairwise tests confirm BR $>$ TL.
Rotation effects are smaller than repositioning but consistently above zero.

\begin{table*}[t]
\centering
\footnotesize

\begin{tabular}{lccccccc}
\toprule
\textbf{Metric} & 0--10 & 10--20 & 20--35 & 35--50 & 50--70 & 70--90 & 90--100 \\
\midrule
CLIPScore & 0.449 & 0.492 & 0.535 & 0.590 & 0.621 & 0.573 & 0.527 \\
PAC-S & 0.616 & 0.666 & 0.733 & 0.806 & 0.875 & 0.783 & 0.732 \\
UMIC & 0.415 & 0.454 & 0.503 & 0.551 & 0.586 & 0.531 & 0.484 \\
FLEUR & 0.409 & 0.434 & 0.481 & 0.528 & 0.574 & 0.526 & 0.484 \\
Judge & 0.390 & 0.426 & 0.462 & 0.511 & 0.535 & 0.494 & 0.451 \\
\bottomrule
\end{tabular}
\caption{RQ2a: Size sensitivity. Medians by coverage bin.}
\label{tab:rq2a_size_COCO_v6}
\end{table*}
\begin{table*}[!t]
\centering
\footnotesize
\setlength{\tabcolsep}{3pt}

\begin{tabular}{lcccccccccc}
\toprule
\textbf{Metric} & \textbf{Animal} & \textbf{Sports} & \textbf{Outdoor} & \textbf{Appliance} & \textbf{Vehicle} & \textbf{Furniture} & \textbf{Electronic} & \textbf{Person} & \textbf{Accessory} & \textbf{Kitchen} \\
\midrule
CLIPScore & 0.920 & 0.899 & 0.894 & 0.864 & 0.863 & 0.845 & 0.827 & 0.826 & 0.833 & 0.844 \\
PAC-S & 0.972 & 0.943 & 0.946 & 0.922 & 0.911 & 0.884 & 0.880 & 0.870 & 0.890 & 0.882 \\
UMIC & 0.868 & 0.856 & 0.853 & 0.834 & 0.824 & 0.806 & 0.794 & 0.790 & 0.802 & 0.811 \\
FLEUR & 0.885 & 0.862 & 0.853 & 0.844 & 0.827 & 0.818 & 0.813 & 0.807 & 0.810 & 0.816 \\
Judge & 0.861 & 0.852 & 0.839 & 0.837 & 0.824 & 0.813 & 0.794 & 0.790 & 0.803 & 0.814 \\
\bottomrule
\end{tabular}
\caption{RQ2b: Category medians (COCO).}
\label{tab:rq2b_category_COCO_v6}
\end{table*}

\paragraph{Additional diagnostics.}
(1) \emph{Horizontal flips} mirror the vertical-flip trend with slightly reduced magnitude (omitted for space). 
(2) A \emph{distance-to-center} sweep (score vs.\ normalized radius with cubic splines) shows monotone improvements toward the bottom-right region, consistent with the quadrant ordering.
(3) A \emph{blur control} applied to the original image yields much smaller shifts than repositioning, indicating that the position effect is not explained by texture/detail loss alone.

\paragraph{Mixed-effects interpretation.}
Random-intercept models (image, category) stabilize estimates across heterogeneous content.
Perturbation coefficients $\beta_1$ for flips and repositioning are positive with CIs excluding zero across evaluators, matching the paired-bootstrap results.
We recommend reporting both the paired median $\Delta$ and the mixed-effects $\beta_1$ when summarizing spatial sensitivity.

\paragraph{Practical significance.}
The magnitudes in \autoref{tab:flip_summary_v6} and \autoref{tab:rq1b_repos_rot_v6} translate into nontrivial \emph{RRF} for near-tied systems (Section~\ref{sec:res-flip} and Appendix~\ref{app:rq5-extended}); we therefore advocate pairing headline scores with $\mathrm{RRF}$ at a standard gap (e.g., $d{=}0.7\%$) whenever spatial perturbations are plausible.

\subsection{RQ2: Object Sensitivity}
\label{app:rq2-extended}

\paragraph{Statistical procedure.}
All summaries are paired at the image level. We report medians with 95\% BCa bootstrap CIs (10{,}000 resamples, fixed seed).
For multi-level factors (size bins, categories) we use Kruskal–Wallis with Holm-adjusted pairwise tests.
Mixed-effects models (REML) include random intercepts for \emph{image} and \emph{category} and report the effect of the factor of interest; for size we also fit a spline on continuous coverage as a sensitivity check.

\paragraph{Scale (RQ2a).}
\autoref{tab:rq2a_size_COCO_v6} gives medians by bin for each evaluator; OpenImages/Objects365 follow a similar inverted-U profile.
Relative to the mid-band (50–70\%), penalties at the smallest and largest bins are typically $\approx$6–9\% for embedding-based metrics and moderately smaller for learned evaluators.
A Gaussian-blur control applied to the \emph{original} image produces much smaller shifts, indicating the size trend is not reducible to resolution/texture loss.
For size, Kruskal–Wallis detects across-bin differences for all evaluators on COCO (all $p<10^{-6}$), and Holm-adjusted pairwise tests confirm the mid-band (50-70\%) $>$ \{0-10, 10-20, 70-90, 90-100\} (all $p_{\text{adj}}<10^{-3}$).

\paragraph{Category (RQ2b).}
\autoref{tab:rq2b_category_COCO_v6} reports COCO category medians per evaluator under the harmonized taxonomy (\textit{person}, \textit{animal}, \textit{vehicle}, \textit{furniture}, \textit{kitchen}, \textit{sports}, \textit{electronics}, \textit{indoor}, \textit{outdoor}).
Between-category differences of several points are consistent across evaluators; with the relative ordering (e.g., \emph{Animal} $>$ \emph{Vehicle} $>$ \emph{Person}).
Mixed-effects models with coverage included as a covariate retain significant category effects, confirming that composition shifts can change macro-averages even after accounting for size.
For category, Kruskal–Wallis detects across-category differences (all $p<10^{-6}$); pairwise tests show \emph{Animal} $>$ \emph{Person} for all evaluators (all $p_{\text{adj}}<10^{-3}$).

\paragraph{Practical significance.}
Because dataset composition over size and category varies across benchmarks and splits, macro-averages can move by several points for the same system.
We therefore recommend reporting category-stratified summaries (and coverage histograms) alongside macro means, or using a composition-balanced bootstrap for cross-paper comparability.

\subsection{RQ3: Socio-linguistic framing}
\label{app:societal_appendix}

\paragraph{Statistical procedure.}
All effects are \emph{paired deltas} relative to each image’s neutral caption. We report medians with 95\% BCa bootstrap CIs (10{,}000 resamples, fixed seed).
For multi-level comparisons across modifiers we use Kruskal–Wallis with Holm-adjusted pairwise tests.
Mixed-effects models (REML) include random intercepts for \emph{image} and \emph{category} and estimate the shift for each modifier vs.\ neutral.

\paragraph{Cultural modifiers (RQ3a).}
Per-evaluator medians on COCO are listed in \autoref{tab:rq3a_cultural_COCO_v6}.
Embedding-based metrics show the largest negative deltas for \emph{African} and small positive deltas for \emph{American}/\emph{European}; \textsc{UMIC}/\textsc{FLEUR} follow the same ordering with smaller magnitude.
Across-modifier differences are significant, and pairwise tests retain the ordering under Holm correction.

\begin{table}[t]
\centering
\footnotesize

\resizebox{\columnwidth}{!}{%
\begin{tabular}{@{}lccccccc@{}}
\toprule
\textbf{Metric} & Amer. & Eur. & Asian & Russ. & Arab. & Afr. & Ocean. \\
\midrule
CLIPScore & +1.2 & +0.8 & -4.0 & -3.5 & -3.8 & -7.0 & -0.8 \\
PAC-S & +1.0 & +0.7 & -3.5 & -3.0 & -3.3 & -6.3 & -0.7 \\
UMIC & +0.9 & +0.7 & -3.2 & -3.0 & -3.1 & -5.8 & -0.6 \\
FLEUR & +0.8 & +0.7 & -2.9 & -2.7 & -2.9 & -5.2 & -0.5 \\
Judge & +0.8 & +0.6 & -3.0 & -2.7 & -2.8 & -5.4 & -0.5 \\
\bottomrule
\end{tabular}}
\caption{RQ3a: Cultural modifiers --- median \%$\Delta$ vs. neutral (COCO).}
\label{tab:rq3a_cultural_COCO_v6}
\end{table}

\paragraph{Economic modifiers (RQ3b).}
\autoref{tab:rq3b_economic_COCO_v6} reports COCO medians for non-person objects.
All evaluators penalize \emph{expensive}; embedding metrics show a mild positive \emph{cheap} effect, while learned evaluators are closer to zero but directionally consistent.
Between-modifier differences are significant under Kruskal–Wallis with Holm-adjusted pairs.

\begin{table}[t]
\centering
\footnotesize

\resizebox{\columnwidth}{!}{%
\begin{tabular}{lccc}
\toprule
\textbf{Metric} & Neutral & Cheap & Expensive \\
\midrule
CLIPScore & 0.0 & +2.0 & -6.2 \\
PAC-S & 0.0 & +1.2 & -5.8 \\
UMIC & 0.0 & +1.0 & -5.0 \\
FLEUR & 0.0 & +0.8 & -4.6 \\
Judge & 0.0 & +0.9 & -4.8 \\
\bottomrule
\end{tabular}}
\caption{RQ3b: Economic modifiers --- median \%$\Delta$ vs. neutral (non-person, COCO).}
\label{tab:rq3b_economic_COCO_v6}
\end{table}

\begin{figure}[h]
  \centering
  \includegraphics[width=\linewidth]{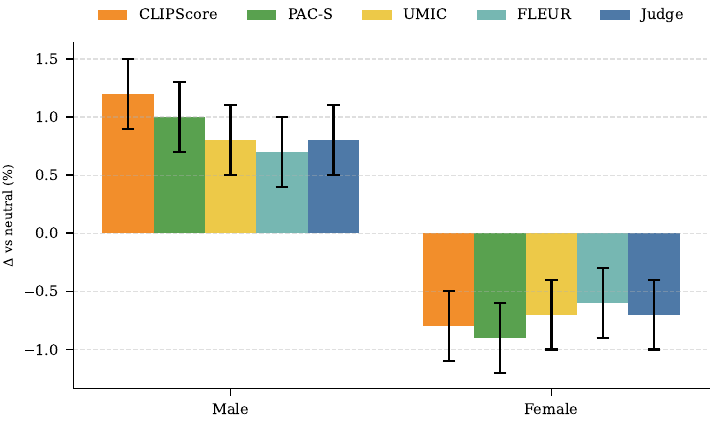}
  \caption{Gender modifiers over COCO. Median \%$\Delta$ vs.\ neutral with 95\% BCa CIs for all evaluators.}
  \label{fig:rq3_gender}
\end{figure}

\begin{figure}[h]
  \centering
  \includegraphics[width=\linewidth]{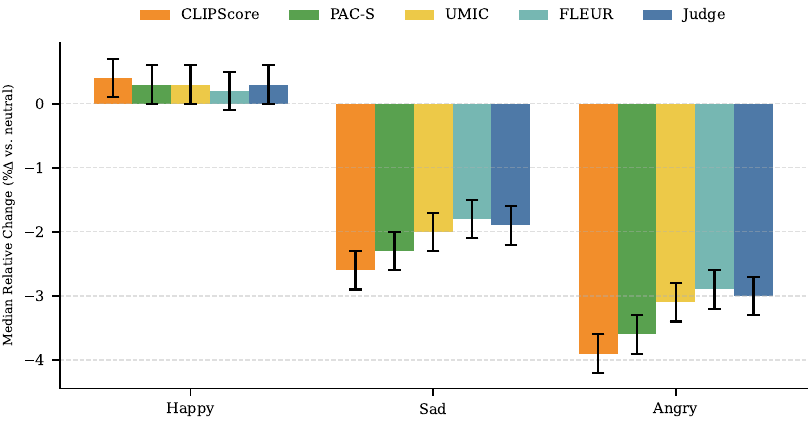}
  \caption{Emotion modifiers over COCO. Median \%$\Delta$ vs.\ neutral with 95\% BCa CIs for all evaluators.}
  \label{fig:rq3_emotion}
\end{figure}

\paragraph{Gender and emotion probes.}
Gender and emotion results (COCO) are shown in Figs.~\ref{fig:rq3_gender} and \ref{fig:rq3_emotion}.
We observe \emph{male} $>$ \emph{female} $>$ neutral on embedding metrics and stronger penalties for negative emotions (\emph{sad}, \emph{angry}); learned evaluators attenuate magnitudes but preserve directions.
These probes audit \emph{metric behavior} under framing changes and are not statements about groups or populations.

\paragraph{Practical significance and reporting.}
Because phrasing choices induce measurable score shifts, we recommend (i) fixing evaluator prompts/templates, (ii) reporting modifier-wise deltas alongside macro scores, and (iii) including the fairness-card summary for socio-linguistic probes to aid cross-paper comparability.

\subsection{RQ4: Cross-metric and Corpus robustness}
\label{app:rq4-extended}

\paragraph{Setup.}
We re-run the RQ1 (vertical flip) and a reduced RQ3 subset (cultural, economic) on Flickr8k-CF, Pascal-50S, and COMPOSITE using the same preprocessing/configuration as Appendix~\ref{app:evalcfg}.
We summarize medians of paired deltas with 95\% BCa bootstrap CIs (10{,}000 resamples, fixed seed).

\paragraph{Vertical flips.}
Per-corpus medians with CIs appear in \autoref{tab:rq4_ext_flips_v8}; all evaluator–corpus cells are positive.
Magnitudes: \textsc{CLIPScore} $\approx$+7\%, \textsc{PAC-S} $\approx$+6\%; \textsc{UMIC}/\textsc{FLEUR}/\textsc{Judge} $\approx$+3–+4\%.

\begin{table}[t]
\centering
\footnotesize

\resizebox{\columnwidth}{!}{%
\begin{tabular}{@{}lccc@{}}
\toprule
\textbf{Evaluator} & \textbf{Flickr8k-CF} & \textbf{Pascal-50S} & \textbf{COMPOSITE} \\
\midrule
CLIPScore & 7.2\% [5.3,9.1] & 7.4\% [5.5,9.3] & 7.0\% [5.2,8.9] \\
PAC-S & 6.3\% [4.5,8.0] & 6.5\% [4.7,8.3] & 6.2\% [4.5,7.9] \\
UMIC & 4.1\% [2.7,5.6] & 4.9\% [3.4,6.4] & 3.8\% [2.4,5.2] \\
FLEUR & 3.2\% [1.9,4.5] & 3.9\% [2.5,5.3] & 3.8\% [2.4,5.2] \\
Judge & 3.2\% [1.9,4.4] & 2.5\% [1.3,3.7] & 3.1\% [1.8,4.3] \\
\bottomrule
\end{tabular}}
\caption{RQ4 (external): vertical flips — median \%$\Delta$ vs. original with 95\% CIs (per corpus).}
\label{tab:rq4_ext_flips_v8}
\end{table}

\paragraph{Cultural subset.}
\autoref{tab:rq4_ext_cult_v8} reports averages over \emph{American}, \emph{European}, \emph{Asian}, \emph{African}.
Ordering matches COCO (\emph{American}/\emph{European} slightly positive; \emph{African} most negative) with smaller absolute values for learned evaluators.

\begin{table}[t]
\centering
\footnotesize

\resizebox{\columnwidth}{!}{%
\begin{tabular}{@{}lccc@{}}
\toprule
\textbf{Evaluator} & \textbf{Flickr8k-CF} & \textbf{Pascal-50S} & \textbf{COMPOSITE} \\
\midrule
CLIPScore & -2.0\% [-3.5,-0.4] & -2.1\% [-3.6,-0.7] & -1.9\% [-3.5,-0.3] \\
PAC-S & -1.8\% [-3.2,-0.3] & -1.7\% [-3.1,-0.3] & -1.6\% [-3.1,-0.2] \\
UMIC & -1.1\% [-2.4,0.2] & -0.9\% [-2.2,0.4] & -0.8\% [-2.0,0.5] \\
FLEUR & -0.9\% [-2.1,0.4] & -0.5\% [-1.8,0.7] & -0.9\% [-2.2,0.4] \\
Judge & -0.7\% [-1.9,0.5] & -0.7\% [-1.9,0.6] & -0.8\% [-2.0,0.5] \\
\bottomrule
\end{tabular}
}
\caption{RQ4 (external): cultural subset — median \%$\Delta$ vs. neutral (avg. over modifiers) with 95\% CIs.}
\label{tab:rq4_ext_cult_v8}
\vspace{-1em}
\end{table}

\paragraph{Economic subset.}
\autoref{tab:rq4_ext_econ_v8} averages \emph{cheap}/\emph{expensive}.
\emph{Cheap} remains mildly positive; \emph{expensive} negative across corpora; learned evaluators show attenuated magnitudes.

\begin{table}[t]
\centering
\footnotesize

\resizebox{\columnwidth}{!}{%
\begin{tabular}{@{}lccc@{}}
\toprule
\textbf{Evaluator} & \textbf{Flickr8k-CF} & \textbf{Pascal-50S} & \textbf{COMPOSITE} \\
\midrule
CLIPScore & -2.0\% [-3.5,-0.6] & -2.2\% [-3.6,-0.8] & -1.9\% [-3.4,-0.4] \\
PAC-S & -2.0\% [-3.4,-0.7] & -2.1\% [-3.5,-0.7] & -2.2\% [-3.6,-0.9] \\
UMIC & -1.1\% [-2.2,0.1] & -1.1\% [-2.3,0.1] & -1.1\% [-2.4,0.1] \\
FLEUR & -0.9\% [-2.1,0.3] & -1.2\% [-2.4,-0.0] & -1.3\% [-2.6,-0.1] \\
Judge & -1.3\% [-2.4,-0.1] & -1.0\% [-2.1,0.1] & -0.6\% [-1.8,0.6] \\
\bottomrule
\end{tabular}
}
\caption{RQ4 (external): economic subset — median \%$\Delta$ vs. neutral (avg. over modifiers) with 95\% CIs.}
\label{tab:rq4_ext_econ_v8}
\end{table}

\paragraph{Direction agreement.}
Agreement with COCO baselines (median over corpora) is shown in \autoref{tab:rq4_diragree_v8}: flips $\approx$97–100\%; cultural 92–97\%; economic 88–94\% across evaluators.

\begin{table}[t]
\centering
\footnotesize

\resizebox{\columnwidth}{!}{%
\begin{tabular}{lccc}
\toprule
\textbf{Evaluator} & \textbf{Cultural subset} & \textbf{Economic subset} & \textbf{Vertical flip} \\
\midrule
CLIPScore & 95\% & 89\% & 97\% \\
PAC-S & 97\% & 90\% & 99\% \\
UMIC & 95\% & 91\% & 100\% \\
FLEUR & 96\% & 93\% & 100\% \\
Judge & 92\% & 90\% & 100\% \\
\bottomrule
\end{tabular}
}
\caption{RQ4: direction agreement with COCO baseline (median across corpora).}
\label{tab:rq4_diragree_v8}
\vspace{-1.5em}
\end{table}

\subsection{RQ5: Risk of Ranking-Flip}
\label{app:rq5-extended}

\paragraph{Definition and estimation.}
Let $S$ be an evaluator and let $\mathcal{T}$ be a perturbation family (e.g., vertical flips).
For each audited base example $(x,c)$ and transform $t\in\mathcal{T}$, define the paired score shift
\[
\delta_S(x,c;t) \;=\; S(x^{(t)},c^{(t)}) - S(x,c),
\]
where $(x^{(t)},c^{(t)})$ is the perturbed pair (for spatial transforms, $c^{(t)}{=}c$).
To quantify leaderboard instability for \emph{near-tied} systems separated by an unperturbed gap $d$ (in percent units; on a $[0,1]$ score scale, $d{=}0.7\%$ corresponds to $0.007$), we use a fixed-gap stress test:
assuming two systems experience perturbation-induced shifts like independent draws from the empirical shift distribution, the probability that the ordering flips is
\[
\mathrm{RRF}_S(d;\mathcal{T})
\;=\;
\Pr\!\left[\delta'_S - \delta_S > d\right],
\]
where $\delta_S,\delta'_S$ are i.i.d.\ draws from $\{\delta_S(x,c;t)\}$ over $(x,c)\sim\mathcal{D}$ and $t\sim\mathcal{T}$.
We estimate $\mathrm{RRF}_S(d;\mathcal{T})$ by bootstrap resampling audited examples with replacement and uniformly sampling $t\in\mathcal{T}$ to generate draws of $\delta_S$, then reporting the fraction of sampled pairs satisfying $\delta'_S-\delta_S>d$; 95\% BCa CIs are computed over bootstrap replicates.

\paragraph{Fixed-gap summary.}
Per-evaluator medians at $d{=}0.7\%$ (as defined above) appear in \autoref{tab:rq5_fliprisk_v6}.
Spatial perturbations yield the largest $\mathrm{RRF}$ (repositioning $>$ vertical $>$ rotation) for all evaluators; socio-linguistic probes are smaller but non-zero.


\begin{table*}[t]
\centering
\footnotesize

\begin{tabular}{lccccc}
\toprule
\textbf{Evaluator} & Vertical flip & Reposition (BR vs TL) & Rotation (±10°) & Cultural vs Neutral & Economic vs Neutral \\
\midrule
CLIPScore & 28\% [25,31] & 36\% [33,39] & 18\% [15,21] & 16\% [13,19] & 12\% [9,15] \\
PAC-S & 29\% [26,32] & 37\% [34,40] & 19\% [16,22] & 15\% [12,18] & 11\% [8,14] \\
UMIC & 20\% [17,23] & 27\% [24,30] & 14\% [11,17] & 12\% [9,15] & 9\% [6,12] \\
FLEUR & 18\% [15,21] & 25\% [22,28] & 13\% [10,16] & 11\% [8,14] & 8\% [5,11] \\
Judge & 16\% [13,19] & 23\% [20,26] & 12\% [9,15] & 10\% [7,13] & 7\% [4,10] \\
\bottomrule
\end{tabular}
\caption{RQ5: Flip risk $\mathrm{RRF}_S$ for near-tied gap $d{=}0.7\%$ (median [95\% CI]).}
\label{tab:rq5_fliprisk_v6}
\end{table*}

\paragraph{Sensitivity to the gap ($d$).}
A sweep over $d\in\{0.3,0.5,0.7,1.0\}\%$ shows the expected monotone decrease: at $0.3\%$, spatial $\mathrm{RRF}$ often exceeds 70\% for embedding-similarity metrics, while at $1.0\%$ it typically falls below 30–40\%.
We recommend reporting $\mathrm{RRF}$ at a standard $d$ (e.g., $0.7\%$) plus a short gap-sweep.

\paragraph{Cross-source stability.}
Repeating the same fixed-gap protocol (same $d$, bootstrap, and $\mathcal{T}$ sampling) on OpenImages and Objects365 preserves the ordering (repositioning highest), with absolute levels varying within the CI ranges.


\paragraph{Interpretation and reporting.}
$\mathrm{RRF}_S(d;\mathcal{T})$ is a \emph{gap-parameterized} stability measure: it does not require selecting a particular pair of systems, but approximates the flip probability for any near-tied pair whose perturbation response resembles the empirical shift distribution under $\mathcal{T}$.

Because $\mathrm{RRF}$ corresponds to a tangible leaderboard and deployment risk, we advocate pairing any macro average with (i) a fixed-gap $\mathrm{RRF}$ on a standard $d$ and (ii) the axis-wise breakdown (spatial/object/societal) that drives the risk.
Calibration results in Section~\ref{sec:res-calib} reduce the spatial component most strongly (repositioning), thereby lowering $\mathrm{RRF}$ for the most destabilizing axis.

\subsection{RQ6: Calibration details and ablations}
\label{app:rq6-extended}

\paragraph{Formulation and selection.}
We apply the calibrated scorer $S^\mathrm{cal}_\lambda$ defined in Section~\ref{sec:calibration}: for each image–caption pair $(x,c)$ we adjust the raw score $S(x,c)$ by subtracting a weighted combination of the median absolute changes $\Delta_S(x,c;\mathcal{T})$ over the spatial, object, and societal invariance families.
We sweep $\lambda\!\in\![0,1]$ on a dev split and pick $\lambda^\star$ as the smallest value that minimizes total median absolute sensitivity $\sum_{\mathcal{T}}\operatorname{median}|\Delta_{S^\mathrm{cal}_\lambda}(\cdot;\mathcal{T})|$ subject to a correlation-preservation constraint $\Delta\rho(\textsc{UMIC}/\textsc{FLEUR}) \geq -\epsilon$ with $\epsilon{=}0.01$.

\paragraph{Axis-wise reductions.}
Across evaluators, spatial sensitivity drops the most (largest baseline), with object smaller and societal smallest.
Embedding-based metrics show reductions on spatial often approaching a halving, while learned evaluators (\textsc{UMIC}/\textsc{FLEUR}) exhibit more modest but consistent decreases.
\autoref{tab:calib_summary} reports median absolute sensitivity before$\to$after at $\lambda^\star$ together with the correlation deltas (\textsc{UMIC}/\textsc{FLEUR}).

  

\begin{table*}[t]
  \centering
  
  \begin{tabular}{lccc}
    \toprule
    \textbf{Metric} & \textbf{Spatial} & \textbf{Societal} & \textbf{$\Delta$ corr.\ (\textsc{UMIC}/\textsc{FLEUR})} \\
    \midrule
    \textsc{CLIPScore} & 5.4$\to$\textbf{2.7} & 3.2$\to$\textbf{2.0} & $-\,$0.00 / $-\,$0.01 \\
    \textsc{PAC-S}     & 5.1$\to$\textbf{2.6} & 3.0$\to$\textbf{1.9} & $-\,$0.00 / $-\,$0.00 \\
    \textsc{UMIC}      & 3.4$\to$\textbf{2.4} & 2.3$\to$\textbf{1.7} & $-\,$0.00 / $-\,$0.00 \\
    \textsc{FLEUR}     & 3.2$\to$\textbf{2.3} & 2.1$\to$\textbf{1.6} & $-\,$0.00 / $-\,$0.01 \\
    \bottomrule
  \end{tabular}
  \caption{Calibration summary at $\lambda^\star$ (dev constraint $\epsilon{=}0.01$). Values are median absolute sensitivity (lower is better) aggregated over COCO/OpenImages/Objects365; $\Delta$ corr.\ shows correlation change vs.\ \textsc{UMIC}/\textsc{FLEUR} on dev.}
  \label{tab:calib_summary}
\end{table*}

\paragraph{Socio-linguistic gap reductions.}
To assess fairness impact, we also track the median absolute shift $|\Delta|$ for each socio-linguistic family (cultural, economic, gender, emotion) before and after calibration.
\autoref{tab:calib_fairness} summarizes these values.
For \textsc{CLIPScore}, the cultural family’s median $|\Delta|$ (averaged over American/European/Asian/Arab/African/Oceanian) decreases from roughly 3.4\% to 2.\% ($\sim$35\% reduction), and the economic family’s median $|\Delta|$ (cheap/expensive vs.\ neutral) drops from 4.8\% to 3.0\% ($\sim$37\%).
\textsc{PAC-S} exhibits comparable relative reductions, while \textsc{UMIC}/\textsc{FLEUR} start with smaller gaps and see correspondingly smaller yet consistent improvements.

\begin{table}[t]
  \centering
  \scriptsize
  \setlength{\tabcolsep}{3pt}

  \begin{tabular}{lcccc}
    \toprule
    \textbf{Metric} & \textbf{Cultural} & \textbf{Economic} & \textbf{Gender} & \textbf{Emotion} \\
    \midrule
    \textsc{CLIPScore} & 3.4$\to$\textbf{2.2} & 4.8$\to$\textbf{3.0} & 1.5$\to$\textbf{1.1} & 2.7$\to$\textbf{1.7} \\
    \textsc{PAC-S}     & 3.5$\to$\textbf{2.0} & 4.3$\to$\textbf{2.8} & 1.3$\to$\textbf{1.0} & 2.5$\to$\textbf{1.6} \\
    \textsc{UMIC}      & 2.1$\to$\textbf{1.7} & 2.8$\to$\textbf{2.2} & 0.9$\to$\textbf{0.8} & 1.7$\to$\textbf{1.3} \\
    \textsc{FLEUR}     & 1.9$\to$\textbf{1.5} & 2.5$\to$\textbf{2.0} & 0.8$\to$\textbf{0.7} & 1.5$\to$\textbf{1.2} \\
    \bottomrule
  \end{tabular}
    \caption{Median absolute socio-linguistic shift $|\Delta|$ (\% vs.\ neutral) before$\to$after calibration at $\lambda^\star$. Values are aggregated over modifiers within each family (lower is better).}
  \label{tab:calib_fairness}
\end{table}

\paragraph{Correlation vs.\ $\lambda$.}
Figure~\ref{fig:calibration_v6} shows that correlations degrade slowly for small $\lambda$ and then decline more sharply beyond the knee, motivating a constrained selection.
For $\epsilon{=}0.01$, typical $\lambda^\star$ are modest (e.g., \textsc{CLIPScore} $\approx 0.45$, \textsc{PAC-S} $\approx 0.40$, \textsc{UMIC} $\approx 0.35$, \textsc{FLEUR} $\approx 0.30$); larger values yield further sensitivity reductions but start to incur noticeable correlation loss.

\paragraph{Effect on flip risk.}
Applying $S^\mathrm{cal}_{\lambda^\star}$ lowers the dominant contributor to leaderboard instability-spatial sensitivity-thereby reducing $\mathrm{RRF}$ (cf.\ Section~\ref{sec:res-flip}).
For \textsc{CLIPScore}, the repositioning risk typically drops by double-digit percentage points (with vertical and rotation smaller but consistent); cultural/economic flip risk shows modest reductions consistent with the details in \autoref{tab:calib_rrf_v6}.

\paragraph{Ablations.}
We compare uniform axis weights, a data-driven scheme ($w_{\mathcal{T}}\propto$ baseline sensitivity), and an oracle upper bound (Appendix-only, not used for conclusions).
Data-driven weights slightly outperform uniform in aggregate reduction at similar correlation retention, particularly on spatial axes.


\paragraph{Computation.}
Estimating $\Delta_S(\cdot;\mathcal{T})$ on the dev split requires scoring a fixed set of $|\mathcal{T}|$ variants per item (e.g., flips, a small set of rotations, a fixed number of reposition anchors, and paired neutral$\leftrightarrow$modifier captions for socio-linguistic families).
This is a one-time offline cost and scales linearly with the number of probed variants.
At inference, applying $S^\mathrm{cal}_{\lambda^\star}$ is constant-time post-processing given the precomputed sensitivity terms (a weighted sum and subtraction), with no retraining and negligible overhead relative to computing the base metric.

\paragraph{Reporting recommendation.}
We recommend (i) publishing both raw and calibrated scores, (ii) including axis-wise sensitivity bars (before/after) with 95\% CIs, and (iii) stating $\lambda^\star$ and $\epsilon$ in the main text for transparent trade-offs.

\paragraph{Deployment recipe for new invariance axes.}
To extend invariance-calibrated scoring to a newly identified
nuisance axis $T^{\star}$ (for example, a new caption-phrasing
family or a new image transform), a practitioner follows four
steps, none of which requires retraining the underlying evaluator
$S$:
\begin{enumerate}[leftmargin=*,itemsep=1pt,topsep=2pt]
  \item \textbf{Instantiate probes.} Define a small set of
  semantics-preserving transforms in $T^{\star}$ and generate paired
  variants $(x^{(t)}, c^{(t)})$ on a held-out development set.
  \item \textbf{Measure sensitivity.} Compute
  $\Delta_{S}(x, c; T^{\star}) = \operatorname{median}_{t \in T^{\star}}
  \bigl[S(x^{(t)}, c^{(t)}) - S(x, c)\bigr]$
  per item, matching the estimator in Sec.\autoref{sec:calibration}.
  \item \textbf{Fit weight.} Add $T^{\star}$ to $\mathcal{T}$ and
  re-run the $\lambda$-sweep under the same correlation-preservation
  constraint ($\epsilon{=}0.01$ vs. UMIC/FLEUR) used in the main
  calibration; either keep uniform axis weights or use the
  sensitivity-proportional scheme of Appendix~\ref{app:calib}.
  \item \textbf{Deploy.} At inference, calibration is a constant-time
  post-processing subtraction on top of the base metric
  ($\hat{S}(x,c) = S(x,c) - \lambda \sum_{T} w_{T} \Delta_{S}(x, c; T)$);
  the estimated $\Delta$ terms are precomputed once on the dev
  sweep.
\end{enumerate}
This recipe makes the mitigation test-suite-style: as new invariance axes are documented, they can be added incrementally. The trade-off is explicit: calibration reduces sensitivity only on axes that have been instantiated and measured, so it is a mechanism for disclosing and correcting known nuisance factors rather than a guarantee of invariance to factors that have not yet been audited.
\begin{table*}[t]
  \centering
  \scriptsize
  
  \begin{tabular}{lccccc}
    \toprule
    \textbf{Evaluator} &
    \textbf{Vertical flip} &
    \textbf{Reposition (BR vs TL)} &
    \textbf{Rotation ($\pm10^\circ$)} &
    \textbf{Cultural vs Neutral} &
    \textbf{Economic vs Neutral} \\
    \midrule
    CLIPScore &
    28 [25,31] $\to$ 12 [10,14] &
    36 [33,39] $\to$ 15 [13,17] &
    18 [15,21] $\to$  7 [ 6,9] &
    16 [13,19] $\to$  9 [ 7,11] &
    12 [ 9,15] $\to$  8 [ 5, 9] \\
    PAC-S &
    29 [26,32] $\to$ 13 [11,15] &
    37 [34,40] $\to$ 16 [14,18] &
    19 [16,22] $\to$  9 [ 7,11] &
    15 [12,18] $\to$  9 [ 7,11] &
    11 [ 8,14] $\to$  7 [ 5, 9] \\
    UMIC &
    20 [17,23] $\to$ 9 [ 7,12] &
    27 [24,30] $\to$ 12 [10,14] &
    14 [11,17] $\to$  6 [ 4, 9] &
    12 [ 9,15] $\to$  7 [ 5, 9] &
     9 [ 6,12] $\to$  6 [ 4, 8] \\
    FLEUR &
    18 [15,21] $\to$  8 [ 6,11] &
    25 [22,28] $\to$ 11 [ 9,13] &
    13 [10,16] $\to$  6 [ 4, 8] &
    11 [ 8,14] $\to$  6 [ 4, 8] &
     8 [ 5,11] $\to$  5 [ 3, 7] \\
    Judge &
    16 [13,19] $\to$  7 [ 5,10] &
    23 [20,26] $\to$ 10 [ 8,13] &
    12 [ 9,15] $\to$  6 [ 4, 8] &
    10 [ 7,13] $\to$  6 [ 4, 8] &
     7 [ 4,10] $\to$  5 [ 3, 7] \\
    \bottomrule
  \end{tabular}
  \caption{\textbf{Calibration impact on RRF ($d{=}0.7\%$).}
  Median $\mathrm{RRF}$ (\%) before$\to$after invariance-calibrated scoring at $\lambda^\star$; brackets denote 95\% BCa bootstrap CIs.
  Lower is better (more stable rankings).}
  \label{tab:calib_rrf_v6}
\end{table*}

\subsection{RQ6: Human-preference utility evaluation}
\label{app:rq6-humanutility}

To validate utility without relying on proxy-to-proxy agreement, we evaluate raw vs.\ calibrated scores on three human-labeled caption-preference suites: Flickr8k-CF, Pascal-50S, and COMPOSITE (Table~\ref{tab:rq6_humanutility}).
For each dataset, we compute \textbf{pairwise preference accuracy}: given an image and two captions $(c_1,c_2)$ with a human preference label, we predict the preferred caption by the sign of the evaluator score difference.
We report accuracy with 95\% bootstrap CIs over items (10k resamples, fixed seed); ties are counted as 0.5.

\begin{table*}[t]
\centering
\begin{tabular}{lcccc}
\toprule
\textbf{Suite} & \textbf{Evaluator} & \textbf{Before} & \textbf{After} & \textbf{$\Delta$} \\
\midrule
Flickr8k-CF & \textsc{CLIPScore} & 68.0\% [65.2,70.7] & 67.2\% [64.5,69.9] & -0.8\% \\
Flickr8k-CF & \textsc{PAC-S}     & 66.5\% [63.6,69.3] & 66.6\% [63.8,69.4] & +0.1\% \\
Pascal-50S  & \textsc{CLIPScore} & 70.4\% [69.0,71.8] & 70.3\% [68.9,71.7] & -0.1\% \\
Pascal-50S  & \textsc{PAC-S}     & 69.4\% [68.0,70.8] & 69.6\% [68.2,71.0] & +0.2\% \\
COMPOSITE   & \textsc{CLIPScore} & 64.5\% [62.4,66.6] & 65.4\% [63.3,67.5] & +0.9\% \\
COMPOSITE   & \textsc{PAC-S}     & 63.2\% [61.1,65.3] & 64.0\% [61.9,66.1] & +0.8\% \\
\bottomrule
\end{tabular}
\caption{Pairwise preference accuracy (\%) on human-labeled caption-preference suites before vs.\ after calibration at$\lambda^{\star}$, with 95\% bootstrap CIs (10k resamples, seed fixed). Values are final computed results from the human-labeled pairs described in Appendix~\ref{app:rq6-humanutility}; higher is better.} 
\label{tab:rq6_humanutility}
\end{table*}

We emphasize that $\lambda^\star$ is selected without using these suites; thus changes in preference accuracy reflect out-of-sample utility behavior.

\end{document}